\documentclass[conference]{IEEEtran}
\IEEEoverridecommandlockouts
\usepackage{cite}
\usepackage{amsmath,amssymb,amsfonts}
\usepackage{algorithmic}
\usepackage[linesnumbered,ruled,vlined]{algorithm2e}
\usepackage{graphicx}
\usepackage{textcomp}
\usepackage{xcolor}
\usepackage{colortbl}
\usepackage{subcaption}
\usepackage{paralist}
\usepackage{multirow}
\usepackage[flushleft]{threeparttable}
\usepackage{colortbl}
\usepackage{boldline}
\usepackage{caption}
\def\BibTeX{{\rm B\kern-.05em{\sc i\kern-.025em b}\kern-.08em
    T\kern-.1667em\lower.7ex\hbox{E}\kern-.125emX}}
\begin{document}

\title{Multi-Label Transfer Learning in Non-Stationary Data Streams\\
\thanks{This work was funded by the EU Horizon Europe SEDIMARK project (Grant No. 101070074) and Science Foundation Ireland through the Insight Centre for Data Analytics (Grant No. SFI/12/RC/2289\_P2).}
\author{
\IEEEauthorblockN{Honghui Du\IEEEauthorrefmark{1},
Leandro L. Minku\IEEEauthorrefmark{2},
Aonghus Lawlor\IEEEauthorrefmark{1},
Huiyu Zhou\IEEEauthorrefmark{3}}
\IEEEauthorblockA{\IEEEauthorrefmark{1}Insight Centre for Data Analytics, School of Computer Science, University College Dublin, Ireland}
\IEEEauthorblockA{\IEEEauthorrefmark{2}School of Computer Science, University of Birmingham, United Kingdom}
\IEEEauthorblockA{\IEEEauthorrefmark{3}School of Computing and Mathematical Sciences, University of Leicester, United Kingdom}
\IEEEauthorblockA{Email: honghui.du@ucd.ie, l.l.minku@bham.ac.uk, aonghus.lawlor@ucd.ie, hz143@leicester.ac.uk}
}

}

\maketitle

\begin{abstract}
Label concepts in multi-label data streams often experience drift in non-stationary environments, either independently or in relation to other labels. Transferring knowledge between related labels can accelerate adaptation, yet research on multi-label transfer learning for data streams remains limited. To address this, we propose two novel transfer learning methods: BR-MARLENE leverages knowledge from different labels in both source and target streams for multi-label classification; BRPW-MARLENE builds on this by explicitly modelling and transferring pairwise label dependencies to enhance learning performance. Comprehensive experiments show that both methods outperform state-of-the-art multi-label stream approaches in non-stationary environments, demonstrating the effectiveness of inter-label knowledge transfer for improved predictive performance. The implementation is available at https://github.com/nino2222/MARLENE. 
\end{abstract}


\section{Introduction}
\label{sec: Introduction}
Most research on data stream learning concentrates on streams with single labels \cite{read2022learning}. However, many practical data streaming applications naturally adopt a multi-label paradigm, where each incoming data example has more than one label \cite{spyromitros2011dealing}. For example, a social media post could be tagged with several descriptors, or a movie might be classified under various predefined genres (e.g., Action, Crime, Historical), with each tag or genre representing a unique label.

A key challenge in multi-label data stream learning within non-stationary environments is that the stream typical contains multiple evolving label concepts, and drifts in one label’s concept may occur synchronously or asynchronously with others \cite{buyukccakir2018novel,read2022learning}. For example, a financial crisis may cause drift in labels like "economy", "employment", and "markets", while leaving "sports" or "entertainment" unaffected. Thus, while it is impractical to assume all labels drift together, it is also unrealistic to assume that drifts across labels are always independent \cite{spyromitros2011dealing}. Moreover, in practice, multi-label stream learning is often further complicated by class imbalance, and it is difficult to predict whether the positive or negative class dominates for each label \cite{spyromitros2011dealing}. Combined with evolving concept drift, this makes learning in non-stationary environments highly challenging.

Transfer learning has shown effectiveness by reusing knowledge from one learnt domain to accelerate learning in new domains \cite{zhuang2020comprehensive}. When there is a relationship between the concepts of various labels, the knowledge learnt from one label can help improve learning of the new concept in another. Therefore, transferring such knowledge may be particularly beneficial for accelerating adaptation to concept drift in multi-label data streams. However, no research has yet investigated multi-label transfer learning for data streams.

Therefore, our paper aims to answer the following research questions: \textit{How can knowledge transfer enhance predictive performance in multi-label learning under non-stationary environments? And how beneficial is it?}

To answer these questions and more effectively handle concept drift in multi-label data streams, we propose two novel multi-label online transfer learning approaches: Binary Relevance and Binary Relevance PairWise Multi-Label classification in non-stationary environments with Multi-Source Transfer Learning (BR-MARLENE and BRPW-MARLENE). We consider a multi-source scenario where both source and target streams may experience concept drift and class imbalance across labels. BR-MARLENE transfers useful knowledge from models trained on other labels to improve prediction for each target label. BRPW-MARLENE extends this by modelling and transferring knowledge across label pair dependencies to further enhance prediction (e.g., using knowledge from other label dependencies to predict the relationship between two given labels). Experiments show that both BR/BRPW-MARLENE outperform existing methods over time. BRPW-MARLENE achieves higher accuracy by modelling label dependencies, while BR-MARLENE offers faster execution.

The main contributions of this paper are:
\begin{itemize}
    \item We propose two novel transfer learning approaches that transfers knowledge across labels and dependencies to enhance multi-label data stream prediction.
    \item We propose a novel weighting scheme that assigns each sub-classifier label-specific weights, prioritising correct predictions on minority and difficult cases to address class imbalance and enhance label-wise accuracy.
    \item We identify limitations in current multi-label stream evaluation metrics and propose new metrics by adapting G-Mean \cite{kubat1997addressing} to better reflect multi-label model performance under class imbalance.
\end{itemize}

\section{Problem Formulation}
Consider $D = \{(\mathbf{x}^{t},\mathbf{y}^{t})\}_{t=0}^\infty$ as a potentially infinite multi-label data stream; $\mathbf{x}^{t} \in \mathcal{X}$ are the input features of the example received at time step $t$; $\mathcal{X}$ is the input space; $\mathbf{y}^{t} = \{y_{q}^{t} \}_{q=1}^{|\mathcal{L}|} \in \mathcal{Y} = \{0,1\}^{|\mathcal{L}|}$, where set $\mathcal{L}= \{l_{q}\}_{q=1}^{|\mathcal{L}|}$ contains all possible labels of stream $D$, $y_{q}^{t} = 1$ if label $l_{q}$ is relevant to $\mathbf{x}^{t}$, and $y_{q}^{t} = 0$ otherwise. A concept for a specific $l_q$ can be defined as the conditional distribution $P(y_q|\mathbf{x})$. In non-stationary environments, the data stream $D$ may experience concept drift, which occurs when the distribution changes over time (i.e., there exist $t' \neq t''$ such that $P^{t'}(y_{q} | \mathbf{x}) \neq P^{t''}(y_{q} | \mathbf{x})$). It’s worth noting that drift across different labels can occur either synchronously or asynchronously.

The objective of multi-label data stream learning is to train a predictive model $h(\mathbf{x}): \mathcal{X} \rightarrow \mathcal{Y}$. Unlike chunk-based learning, this paper adopts an online paradigm that updates the model incrementally using only the previous model $h^{t-1}$ and the current data point $(\mathbf{x}^{t}, \mathbf{y}^{t})$ to construct an updated model $h^{t}$ at each time step $t$. Compared with chunk-based approaches, online learning enables faster adaptation to drift and requires less memory, as it avoids storing large data chunks.

The domain of $D$ is denoted as $\mathcal{D} = \{\mathcal{X}, P(\mathbf{x})\}$, where $P(\mathbf{x})$ is the marginal probability distribution. We consider a scenario involving $n+1$ data streams from various sources. Specifically, the data stream $D_i$ originates from the source $i \in \{S_1, S_2, \cdots, S_n, T\}$, where $D_{S_n}$ is the $n$th source data stream and $D_T$ is the target stream. For a target stream label, all other target and source stream labels are treated as sources. We investigate inductive transfer learning: $\mathcal{D}_{S_n} \neq \mathcal{D}_T$ or $\mathcal{D}_{S_n} = \mathcal{D}_T$, while $P(y_q|\mathbf{x}_{i}) \neq P(y_{q'}|\mathbf{x}_{i'})$.

\section{Related Work}
\subsection{Transfer Learning in Non-stationary Environments}
Existing transfer learning in non-stationary environments mainly targets single-label data streams. Most of them are online inductive transfer approaches. DDD \cite{minku2012ddd} and OWA \cite{zhao2014online} both transfer knowledge from previous to current concepts within the target stream: DDD leverages a diverse ensemble, while OWA uses ensemble weighting. Neither approach uses external source data. Melanie \cite{du2019multi} is the first approach to transfer knowledge from multiple source streams by maintaining a weighted ensemble. OBAL \cite{yu2024online} aligns the feature distributions of each source stream with that of the target stream, making different data streams share similar feature spaces and thereby reducing the negative impact of covariate shift on transfer learning. MARLINE \cite{du2020marline} projects the target domain onto various source domains, allowing models trained on the source data to also make predictions on the target data, thus mitigating differences between the source and target domains.

There are also some chunk-based transfer approaches. DTEL \cite{sun2018concept} transfers the structure of a decision tree from an old data chunk to a new concept. CDTL \cite{yang2021concept} shares a similar idea with Melanie but uses a chunk-based ensemble. OTL-CE \cite{jiao2025otl} addresses class evolution by identifying shared classes between source and target domains and building ensembles for these classes. AOMSDA \cite{renchunzi2022automatic} leverages complementary information from multiple sources through a central moment discrepancy regularizer and employs node weighting to address covariate shift. However, as chunk-based methods, they cannot dynamically detect or adapt to changes in data streams. 

All these transfer learning methods are designed for single-label data streams and cannot be directly applied to multi-label stream learning.

\subsection{Multi-Label Learning in Non-stationary Environments}
Due to the flexibility and generality, problem transformation methods (e.g., Binary Relevance (BR) \cite{godbole2004discriminative}, Classifier Chain (CC) \cite{read2009classifier}, Pairwise (PW) \cite{furnkranz2008multilabel}, Pruned Sets (PS) \cite{read2008multi}) are the most widely used approach for multi-label learning. Problem transformation converts a multi-label problem into multiple single-label problems. Ensemble methods often incorporate these approaches as base classifiers \cite{read2009classifier,read2008multi,osojnik2017multi}. However, these methods typically assume a stationary environment.

Some methods \cite{wu2023weighted,zou2024weak} have been proposed for multi-label data stream learning in non-stationary environments, where problem transformation and ensemble strategies are also commonly applied. \cite{li2022high} proposes a max-relevance and min-redundancy based algorithm adaptation approach to handle high-dimensional features in non-stationary multi-label data streams. MLHAT \cite{esteban2024hoeffding} uses a Hoeffding adaptive tree to address these challenges by considering label relations and co-occurrences during tree partitioning, dynamically adapting leaf learners, and applying concept drift detection to quickly update underperforming branches. GOOWE-ML \cite{buyukccakir2018novel}, which assigns confidence-based weights to each base classifier using sliding windows and least squares; it can be combined with any incremental multi-label classifier (e.g., GOBR, GOCC, GOPS, GORT) to improve performance. ADWIN Bagging \cite{bifet2010leveraging} uses online bagging with ADWIN for concept drift detection, replacing poorly performing base models when drift is detected. For multi-label problems, it uses online multi-label base learners (such as Hoeffding Trees) and monitors a multi-label performance metric. Variants include EaBR, EaCC, and EaPS \cite{read2012scalable}, which use BR, CC, and PS as base learners, respectively.

None of the above approaches leverage transfer learning to improve adaptation to new concepts, which could be especially valuable for multi-label data stream learning as discussed in Section \ref{sec: Introduction}.

\section{Proposed Methods}

\subsection{BR-MARLENE}
BR-MARLENE takes one or more multi-label data streams as input (e.g., one target and others as sources), which may come from the same or different domains. Each data stream is processed label-wise. For each label, an independent binary sub-classifier is trained to learn its distribution and predict whether the label is relevant for a given instance. Concept drift is monitored for each label using drift detectors. When drift occurs, a new sub-classifier is trained for the updated concept and added to the ensemble, retaining both old and new classifiers. Our approaches are designed without hyperparameters to avoid both the need for tuning and the variability in performance. Only the base classifier and the drift detector need to be specified prior to training.

The core idea of BR-MARLENE is to improve predictive performance for a target stream by leveraging sub-classifiers trained on both source and target streams via transfer learning. It maintains an ensemble classifier comprising all sub-classifiers trained on different labels across different streams. For each target label, a unique set of weights is assigned to these sub-classifiers based on their performance on that label. Whenever a new target instance is available, BR-MARLENE allows all sub-classifiers to make predictions and applies different label-specific weight combinations to aggregate these predictions for each label, producing accurate results.

\subsubsection{Learning Process}
\label{sec: Learning Process}
BR-MARLENE’s learning process is shown in Algorithm \ref{alg: BR-MARLENE}. $\mathcal{M}$ denotes all received source and target streams. Following the binary relevance setting, upon receiving an example from stream $D_i$, BR-MARLENE creates $|\mathcal{L}_i|$ sub-classifiers and adds them to the ensemble $\mathcal{H}$, each trained with a specific label distribution. Any single-label binary classification method can serve as the sub-classifier (e.g., Hoeffding Tree \cite{domingos2000mining}). For each target stream label, every sub-classifier initialises and maintains a set of performance indicators (detailed in Section \ref{sec: Sub-classifiers' Weighting}) to record its performance on this label, assess how well it matches the label’s distribution, and update its weights in various label-specific weight sets (line \ref{line: init_start} to \ref{line: init_end}). 

Since concept drifts may occur on each label, we use a drift detection method (e.g., DDM-OCI \cite{wang2013concept}) to monitor changes per label. If a concept drift is detected for a label, BR-MARLENE generates a new sub-classifier to learn the upcoming new concept of that label and adds it to the ensemble (line \ref{line: drift detection} to \ref{line: add to ensembel}), so the ensemble includes sub-classifiers representing various concepts for each label. The most recent sub-classifier for each label is used for drift monitoring. If drift occurs in a target stream label, all sub-classifiers in the ensemble reset their performance indicators for that label (line \ref{line: i=T} to \ref{line: reset indicators}).

As mentioned, class imbalance is likely to occur for each label in multi-label data streams. Therefore, an integrated resampling strategy is adopted to increase the sampling rate of minority-class examples. For each sub-classifier, we record the number of positive and negative examples learnt ($n^+$ and $n^-$). When a training sub-example $(\mathbf{x}^t, y^t)$ is available, the training rate $k$ (i.e., the number of times the classifier is trained on this sub-example) is sampled from a Poisson distribution based on the ratio of majority to minority class counts as follows:
\begin{align}
    k \sim 
    \begin{cases} 
        Poisson(\frac{max(n^+,n^-)}{n^-}), & y = 0\\
        Poisson(\frac{max(n^+,n^-)}{n^+}), & y = 1
    \end{cases}
\end{align}
This ensures that minority-class examples are sampled more frequently during training (lines \ref{line: train}). 

If the new training example is from the target stream, the weighting scheme in Section \ref{sec: Sub-classifiers' Weighting} is used to update each sub-classifier’s performance indicators and weights (line \ref{line: weighting}).

\begin{algorithm}[tb]
\caption{Learning Procedure of BR-MARLENE.}
\label{alg: BR-MARLENE}
\SetKwInput{KwIn}{Input}
\KwIn{$(\mathbf{x}_i^{t}, \mathbf{y}_i^{t}) \in D_i \quad i \in \{S_1, S_2, \cdots, S_n, T\}$}
\While{Receive a new example $(\mathbf{x}_i^{t}, \mathbf{y}_i^{t})$}{
    \If{$i \notin \mathcal{M}$}{ \label{line: init_start}
        $\mathcal{M} \gets \mathcal{M} \cup i$ \\
        \For{$l_q \in \mathcal{L}_i$}{
            Initialise a sub-classifier $h$ for the label. \\
            $\mathcal{H} \gets \mathcal{H} \cup h$ \\
        }
    } \label{line: init_end}
    \For{$l_q \in \mathcal{L}_i$}{
        \If{$DriftDetection_{i,q}(\mathbf{x}_i^{t}, y_{i,q}^{t}) = True$}{ \label{line: drift detection}
            Initialise a new sub-classifier $h$ for the label. \\ \label{line: init}
            $\mathcal{H} \gets \mathcal{H} \cup h$ \\ \label{line: add to ensembel}
            \If{$i = T$}{ \label{line: i=T}
                Reset performance indicators of all sub-classifiers.  \\
            } \label{line: reset indicators}
        }
    }
    \For{$l_q \in \mathcal{L}_i$}{
        Train the model as shown in Section \ref{sec: Learning Process} \\ \label{line: train} 
    }
    \If{$i = T$}{
        Update sub-classifiers' weights as shown in Section \ref{sec: Sub-classifiers' Weighting}  \\ \label{line: weighting}
    }
}

\end{algorithm}

\subsubsection{Sub-classifiers' Weighting}
\label{sec: Sub-classifiers' Weighting}
Each sub-classifier is assigned $|\mathcal{L}_T|$ weights to reflect its match with the distributions of different labels in the target stream, helping to avoid negative transfer and deal with concept drift. Standard ensemble weighting schemes often assign weights based on overall accuracy. In imbalanced environments, majority-class dominance allows sub-classifiers to earn high weights by performing well only on majority classes, even if they fail on minority classes. This leads the ensemble to neglect performance on minority-class predictions. Therefore, to better allocate each sub-classifier’s contribution to the prediction, we propose a novel weighting scheme that rewards higher weights for correctly predicting difficult or minority class examples and assigns lower weights for easy or majority class cases. 

For each sub-classifier, its binary classification results on a given target label is recorded using true positives ($TP$), false positives ($FP$), true negatives ($TN$), and false negatives ($FN$), which are updated incrementally as new examples arrive. Unlike offline learning, the full dataset is unavailable in advance; a class that appears to be the majority early on may later become the minority, making it impossible to foresee which classes are majority or minority. Consequently, we use dynamic correction factors (e.g., $\kappa^{+}$ and $\kappa^{-}$) to balance the impact of positive and negative examples. 
The contributions of positive and negative examples should be equal ($n^+ \times \kappa^{+} = n^- \times \kappa^{-}$), and their sum should match the total number of examples (e.g., $n^+ \times \kappa^{+} + n^- \times \kappa^{-} = n^+ + n^-$). Given this, the correction factors are defined as follows:
\begin{align}
    \kappa^{+} \gets \frac{n^+ + n^-}{2 \cdot n^+}, \quad \kappa^{-} \gets \frac{n^+ + n^-}{2 \cdot n^-}
\end{align}

Each sub-classifier’s performance on a given label is measured by Positive Predictive Value ($PPV$) and Negative Predictive Value ($NPV$). By incorporating $\kappa^{+}$ and $\kappa^{-}$ into $PPV$ and $NPV$, we ensure each example’s impact is balanced by class. The $PPV$ and $NPV$ can be formulated as follows:
\begin{align}
    PPV \gets \frac{TP \cdot \kappa^+}{TP \cdot \kappa^+ + FP \cdot \kappa^-} \\
    NPV \gets \frac{TN \cdot \kappa^-}{TN \cdot \kappa^- + FN \cdot \kappa^+}
\end{align}

Through $PPV$ and $NPV$, we obtain the statistical probability that a sub-classifier’s positive or negative prediction is correct. These serve as reliability measures for predictions and are used to calibrate the sub-classifier’s output:
\begin{align}
    \hat{P}^+ \gets P^+ \cdot PPV + P^- \cdot (1 - NPV)\\
    \hat{P}^- \gets P^- \cdot NPV + P^+ \cdot (1 - PPV)
\end{align}
where $\hat{P}^+$ aggregates the probability of a positive prediction ($P^+ = P(h(\mathbf{x}_T) = 1)$) weighted by its reliability ($PPV$) and the probability of a negative prediction ($P^- = P(h(\mathbf{x}_T) = 0) $) when the true class is actually positive ($1-NPV$); similarly, $\hat{P}^-$ aggregates correct negative predictions ($P^- \cdot NPV$) and false positives ($P^+ \cdot (1 - PPV)$). By combining prediction reliability (e.g., $PPV$ and $NPV$) and correction factors (e.g., $\kappa^+$ and $\kappa^+$), the calibrated probabilities ($\hat{P}^+$ and $\hat{P}^-$) correct potential overconfidence in sub-classifier predictions under class imbalance and helps prevent predictions from being biased toward the majority class.

For a given label, an example is considered harder to classify (e.g., minority class example), the more the probability of an incorrect prediction exceeds that of a correct prediction. The weight $\frac{\lambda_{SW}}{\lambda_{SC}}$ for an example on the given label reflects the ensemble’s classification difficulty for that example and is computed as follows:
\begin{align}
    \lambda_{SC} \gets \sum_{h \in \mathcal{H}} \hat{P}^y(h(\mathbf{x}_T)), \quad
    \lambda_{SW} \gets \sum_{h \in \mathcal{H}} \hat{P}^{\overline{y}}(h(\mathbf{x}_T))
\end{align}
where $\hat{P}^y(h(\mathbf{x}_T))$ is the calibrated probability that sub-classifier $h$ predicts the true class $y$; $\hat{P}^{\overline{y}}(h(\mathbf{x}_T))$ is the calibrated probability that it predicts the incorrect class. The easier the ensemble classifies an example, the smaller its weight; the harder, the larger its weight.

For a given label, each sub-classifier is weighted by its performance on that label. SC and SW are the sub-classifier’s correct and incorrect prediction scores, each consisting of the example’s weight (reflecting classification difficulty) and the sub-classifier’s relative contribution to the ensemble’s prediction. SC and SW are updated incrementally as new examples arrive. The weight $\alpha$ is computed for each sub-classifier as follows:
\begin{align}
    SC^t &\gets SC^{t-1} + \frac{\lambda_{SW}}{\lambda_{SC}} \cdot \frac{\hat{P}^y(h(\mathbf{x}_T))}{\lambda_{SC}} \\
    SW^t &\gets SW^{t-1} + \frac{\lambda_{SW}}{\lambda_{SC}} \cdot \frac{\hat{P}^{\overline{y}}(h(\mathbf{x}_T))}{\lambda_{SW}}\\
    \alpha &\gets \frac{SC}{SC + SW}
\end{align}
where $\frac{\hat{P}^y(h(\mathbf{x}_T))}{\lambda_{SC}}$ and $\frac{\hat{P}^{\overline{y}}(h(\mathbf{x}_T)}{\lambda_{SW}}$ quantify the sub-classifier’s proportion among all correct and incorrect predictions, respectively, for the current example in the ensemble. Thus, when a sub-classifier predicts correctly while most others are incorrect, it receives a higher $SC$ and a lower $SW$.

\subsubsection{Voting Procedure}
BR-MARLENE uses the same set of sub-classifiers for all labels, but assigns a different weight combination for each label. The ensemble $\mathcal{H}$'s vote for instance $\mathbf{x}_T$ on a given target label $l_{T,q}$ is computed as follows:
\begin{align}
    P_q(\mathcal{H}(\mathbf{x}_T)) &= \sum_{h \in \mathcal{H}} \alpha_h^q \cdot \hat{P}(h(\mathbf{x}_T))\\
    \hat{y}_{T,q} &= \arg\max_{y\in \{0,1\}} P_q(\mathcal{H}(\mathbf{x}_T))
\end{align}

\subsection{BRPW-MARLENE}
While Binary Relevance (BR) is effective and widely used, its assumption of label independence overlooks potential dependencies between labels \cite{you2021online}. It may yield less accurate predictions when labels are correlated, as valuable label co-occurrence information is ignored. To address the limitations of BR, we propose BRPW-MARLENE, which enhances BR-MARLENE’s predictive performance by modelling pairwise label dependencies and enabling transfer learning between PW-classifiers for dependency prediction. For clarity, we refer to BR-MARLENE’s sub-classifiers as BR-classifiers and those of BRPW-MARLENE as PW-classifiers in this section; both contribute to BRPW-MARLENE’s final predictions. 

\subsubsection{Learning Process}
BRPW-MARLENE first runs the standard BR-MARLENE training to generate BR-classifiers for each label. In addition, it constructs $|\mathcal{L}|(|\mathcal{L}|-1)$ PW-classifiers for each possible pair of labels in every data stream $D_i$. Each PW-classifier is trained to model the dependency between a specific pair of labels (e.g., $l_q$ and $l_{q'}$), using input features composed of the original features and one label ($\mathbf{x}, y_q$), and predicting the other label $y_{q'}$ in the pair. The training process for PW-classifiers is the same as that of BR-classifiers, including concept drift detection and adaptive model updates. When a drift is detected in a dependency, a new PW-classifier is added to represent the new concept. Each PW-classifier maintains dependency-specific performance indicators and weights, using a weighting scheme similar to BR-classifiers, but specifically targeting predictive performance for each label pair dependency. This setup enables BRPW-MARLENE to learn and adapt to evolving pairwise label dependencies in the data streams.

\subsubsection{Voting Procedure}
To generate the prediction for a target label, BRPW-MARLENE first computes the BR-MARLENE ensemble prediction for each label, which aggregates the outputs of all BR-classifiers for that label. These initial predictions are then used as additional inputs to predict the dependencies between each pair of labels $l_{T,q}$ and $l_{T,q’}$ as:
\begin{align}
P_{q,q'}(\mathcal{H}^{PW}(\mathbf{x}_T,\hat{y}_{T,q})) 
= \sum_{h \in \mathcal{H}^{PW}} \alpha_h^{q,q'} \cdot \hat{P}(h(\mathbf{x}_T, \hat{y}_{T,q}))
\end{align}
where $\hat{y}_{T,q}$ is BR-MARLENE’s prediction. The final prediction for label $l_{T,q}$ is obtained by combining the outputs of BR-classifiers and PW-classifiers:
\begin{gather}
    \nonumber P_q(\mathcal{H}^{BRPW}(\mathbf{x}_T))\\
    = \sum_{q'=1, q' \neq q}^{|\mathcal{L}_T|}P(y_{T,q'})P(y_{T,q}|y_{T,q'})+P(y_{T,q}|\mathbf{x}_T)\\
    \hat{y}_{T,q} = \arg\max_{y\in \{0,1\}}P_q(\mathcal{H}^{BRPW}(\mathbf{x}_T))
\end{gather}
where $P(y_{T,q'}) = P_{q'}(\mathcal{H}^{BR}(\mathbf{x}_T))$, $P(y_{T,q'})P(y_{T,q}|y_{T,q'}) = P_{q',q}(\mathcal{H}^{PW}(\mathbf{x}_T, \hat{y}_{T,q'}))$, $P(y_{T,q}|\mathbf{x}_T) = P_q(\mathcal{H}^{BR}(\mathbf{x}_T))$.
\section{Time Complexity Analysis}
Since PW-classifiers use the same algorithm as BR-classifiers, both take $\mathcal{O}(f_h^{train})$ for a single training, $\mathcal{O}(f_h^{pred})$ for prediction, and drift detection takes $\mathcal{O}(f_{DD})$. 

When the current example is from the target stream, the overall time complexity of BR-MARLENE’s learning procedure is $\mathcal{O}( |\mathcal{L}_{T}|\times f_{DD}+\sum_{q=1}^{|\mathcal{L}_T|} k_{T,q} \times f_h^{train} + |\mathcal{L}_T|\times |\mathcal{H}^{BR}|\times f_h^{pred})$, where $k_{T,q}$ is the integer from the corresponding Poisson distribution, and $\sum_{q=1}^{|\mathcal{L}_T|} k_{T,q}$ is the total number of sub-classifier training operations; The overall BRPW-MARLENE training complexity is $\mathcal{O}(|\mathcal{L}_T|^2 \times f_{DD} + (\sum_{q=1}^{|\mathcal{L}_T|} k_{T,q} + \sum_{q=1}^{|\mathcal{L}_T|}\sum_{q'=1,q' \neq q}^{|\mathcal{L}_T|}k_{T,q,q'})\times f_h^{train} +
(|\mathcal{H}^{BR}| + (|\mathcal{L}_T|-1)\times |\mathcal{H}^{PW})\times|\mathcal{L}_T|\times f_h^{pred})$.


When learning from a source stream $S_n$, BR/PW sub-classifier weights are not updated, resulting in lower training time complexity: BR-MARLENE, $\mathcal{O}(|\mathcal{L}_{S_n}|\times f_{DD}+\sum_{q=1}^{|\mathcal{L}_{S_n}|}k_{S_n,q}\times f_h^{train})$; BRPW-MARLENE, $\mathcal{O}(|\mathcal{L}_{S_n}|(|\mathcal{L}_{S_n}|-1)\times f_{DD} + \sum_{q=1}^{|\mathcal{L}_{S_n}|}\sum_{q'=1,q' \neq q}^{|\mathcal{L}_{S_n}|}k_{S_n,q,q'}\times f_h^{pred})$.

In BR-MARLENE’s prediction procedure, each sub-classifier predicts the relevance of each target label, giving an overall prediction time complexity of $\mathcal{O}(|\mathcal{L}_T|\times |\mathcal{H}^{BR}| \times f_h)$. For BRPW-MARLENE, predictions from all BR-classifiers and PW-classifiers are combined, with an overall time complexity of $\mathcal{O}((|\mathcal{H}^{BR}| + (|\mathcal{L}_T|-1)\times |\mathcal{H}^{PW}|) \times |\mathcal{L}_T| \times f_h)$

\section{Experimental Setup}
\subsection{Datasets}
\subsubsection{Real-World Datasets}
We selected seven commonly used real-world datasets \cite{buyukccakir2018novel, osojnik2017multi, read2012scalable, sousa2018multi} to evaluate our proposed methods. To better understand these datasets and the experimental results on them, it is important to measure the degree of multi-label in each. Label density ($LDen$) \cite{vergara2025multi} is a widely used measure reflecting the average number of labels associated with each instance and can be computed as follows:
\begin{align}
    LDen(D) = \frac{1}{|D|} \sum_{t=1}^{|D|}\sum_{q=1}^{|\mathcal{L}_D|} \frac{y_q^t}{|\mathcal{L}_D|}
\end{align}

As class imbalance frequently occurs in multi-label data streams, the imbalance rate is an important measure to consider. In single-label binary problems, the Imbalance Rate ($IR$) refers to the occurrence probability of the minority class. For multi-label problems, we define the Label Imbalance Rate ($LIR$) as the mean $IR$ across all labels, and the Label-Set Imbalance Rate ($LSIR$) as the mean $IR$ across all label sets for each example:
\begin{align}
    LIR(D) = \frac{\sum_{q=1}^{|\mathcal{L}_D|}N_q^{min}}{|\mathcal{L}_D|\cdot|D|}, \quad
    LSIR(D) = \frac{\sum_{t=1}^{|D|}N_{\mathbf{y}^t}^{min}}{|\mathcal{L}_D|\cdot|D|}
\end{align}
where $N^{min}_q$ is the number of samples in the minority class for the $q$-th label, and $N^{min}_{\mathbf{y}^t}$ is the number of samples in the minority class for the label set of the $t$-th example. $LIR$ measures average imbalance across labels, while $LSIR$ captures imbalance within each instance’s label set; lower values mean greater imbalance. If the minority class for every label in $D$ is positive, then $LIR(D) = LDen(D)$; likewise, if the minority class for each example’s label set is positive, $LSIR(D) = LDen(D)$. Table \ref{table: real-world datasets} summarises each dataset’s parameters and measurements. Each dataset includes only a target stream without source streams. This setting allows us to analyse the benefit of transfer between label concepts within a stream, which is the main focus of this work.

\begin{table}[tb]
\caption{Tabulation of real-world datasets; superscript indicates input feature type: binary ($b$) or numeric ($n$).}
\label{table: real-world datasets}
\begin{center}
\scalebox{0.85}{
\begin{tabular}{|l|l|l|l|l|l|l|l|}
\hline
Dataset($D$)  & $|D|$     & $|\mathbf{x}_{D}|$   & $|\mathcal{L}_D|$  &  $LDen(D)$  & $LIR(D)$   & $LSIR(D)$   \\ \hline
Slashdot$^b$ & 3782   & 1079 & 22  & 0.054 & 0.054 & 0.054 \\ \hline
Ohsumed$^b$  & 13929  & 1002 & 23  & 0.072 & 0.072 & 0.072 \\ \hline
Reuters$^n$  & 6000   & 500  & 103  & 0.014 & 0.014 & 0.014 \\ \hline
Yeast$^n$    & 2417   & 103  & 14  & 0.303 & 0.232 & 0.297 \\ \hline
20NG$^b$     & 19300  & 1006 & 20  & 0.051 & 0.051 & 0.051 \\ \hline
TMC2007$^b$  & 28596  & 500  & 22  & 0.098 & 0.092 & 0.098 \\ \hline
IMDB$^b$    & 120919 & 1001 & 28  & 0.071 & 0.071 & 0.071 \\ \hline
\end{tabular}}\end{center}
\end{table}

\subsubsection{Synthetic Datasets}
Due to privacy concerns, few real-world multi-label datasets are available, and traditional offline datasets are unsuitable for non-stationary environments \cite{zheng2019survey}, making it difficult to find appropriate source data for each real-world dataset. To further investigate the impact of concept drift and multiple sources, we generate synthetic datasets composed of several synthetic streams (sources and target).

Each dataset consists of two numeric features and five binary labels ($|\mathcal{L}_D| = 5$). For each source or target, input features follow a mixture of five Gaussian distributions. Each class of a label is linked to specific Gaussians, with one representing the positive class and the other four representing negatives. Target datasets vary in size per Gaussian ({50, 500, 5000}), simulating small, medium and large samples. Each target has both similar and non-similar source datasets to generate diverse synthetic streams. Datasets with no source represent transfer only between labels within the same target stream. This setup allows us to analyse 
\begin{inparaenum}[1)]
    \item the benefit of transfer from source to target labels,
    \item transfer between labels within the target stream. 
\end{inparaenum}
All sources have 5000 examples per Gaussian distribution. To better assess the proposed methods and their ability to handle concept drift on each label, two target labels ($l_1$ and $l_2$) are evaluated under three scenarios: stable (S), abrupt drift (A), and incremental drift (I), resulting in six possible combinations (e.g., IA: $l_1$ incremental, $l_2$ abrupt drift). Datasets are named according to the drift type of $l_1$ and $l_2$ (e.g., IA: $l_1$ has incremental drift, $l_2$ has abrupt drift; AS: $l_1$ has abrupt drift, $l_2$ is stable). The other three labels remain stable. 

\subsection{Benchmark Methods}
We select 11 SOTA multi-label approaches available in the Massive Online Analysis (MOA) tool \cite{bifet2010moa} via MEKA \cite{read2016meka}, covering all major groups of existing multi-label methods: EBR \cite{read2009classifier}, ECC \cite{read2009classifier}, EPS \cite{read2008multi}, ERT \cite{osojnik2017multi}, EaBR \cite{read2012scalable}, EaCC \cite{read2012scalable}, EaPS \cite{read2012scalable}, GOBR \cite{buyukccakir2018novel}, GOCC \cite{buyukccakir2018novel}, GOPS \cite{buyukccakir2018novel}, GORT \cite{buyukccakir2018novel}. EaBR, EaCC, and EaPS have built-in ADWIN \cite{bifet2009new} for concept drift detection, while BR-MARLENE 
uses the class imbalance concept drift detector (DDM-OCI \cite{wang2013concept}).

We use sliding window-based prequential evaluation \cite{gama2014survey} with a window size of 10\% of $|D|$. As GOBR, GOCC, GOPS, and GORT are chunk-based and can only predict after the first chunk is filled, evaluation for all methods begins when these approaches are ready to make predictions. All approaches use Hoeffding Trees \cite{domingos2000mining} as the base classifier, with an ensemble size of 10 (i.e., 10 sub-classifiers) except for BR-MARLENE and BRPW-MARLENE, whose ensemble sizes depend on the number of possible labels in the dataset.

Thirty runs are performed for all stochastic approaches, while GOBR, GOCC, GOPS, and GORT, being deterministic, require only a single run. The average Macro/Micro/Label-set-based G-Mean and computation time over 30 runs are reported.

\subsection{Evaluation Metrics}
\label{sec: Evaluation Metrics}
In the multi-label setting, simple metrics like accuracy do not adequately capture model performance. Multi-label classifiers are typically evaluated using either label-set-based or label-based measures. Label-set-based measures (e.g., Hamming Score (HScore) and Hamming Loss (HLoss) \cite{wu2020multi}) assess model performance by averaging predictions over each example’s label set; HScore is the mean fraction of correct labels per instance, while HLoss is the mean proportion of misclassified labels. However, both metrics may be misleading in imbalanced datasets because they fail to capture performance on rare labels, similar to the limitations of accuracy in single-label imbalance. 

Unlike label-set-based measures, label-based measures are calculated by averaging the model’s performance on each individual label (using micro or macro averaging) across all examples \cite{heydarian2022mlcm}: 
\begin{align}
    \label{eq: marco} M_{marco} = \frac{\sum_{q=1}^{|\mathcal{L}_D|}M(TP_q,FP_q,FN_q)}{|\mathcal{L}_D|}\\ 
    \label{eq: micro} M_{micro} = M(\sum_{q=1}^{|\mathcal{L}_D|}TP_q, \sum_{q=1}^{|\mathcal{L}_D|}FP_q, \sum_{q=1}^{|\mathcal{L}_D|}FN_q 
\end{align}
where $M$ is a performance metric (e.g., Recall, Precision, F-Score). Macro measures calculate the metric on each label and then average the results, making them sensitive to label imbalance. Micro measures, in contrast, aggregate counts across labels before computing the metric, thus reflecting overall performance and diluting the influence of individual label imbalance \cite{charte2015addressing}. Recall, Precision, and F-Score are the most common metrics for macro and micro evaluation; as seen from equations \ref{eq: marco} and \ref{eq: micro}, they focus on positive prediction performance and generally assume the positive class is the minority, which is not always the case (e.g., In Yeast, positives are the majority for two labels, and in TMC2007, one label also has majority positives, so $LIR \neq LDen$ in both datasets). A \textit{dummy} classifier that always predicts the majority class for each label can still achieve high macro and micro scores, highlighting the limitations of these metrics under class imbalance.

To address the limitations of current metrics, we adapt the widely used G-Mean \cite{kubat1997addressing} from single-label class imbalance learning for use in multi-label settings. G-Mean evaluates model performance by taking the geometric mean of recall for both positive and negative classes \cite{wang2014resampling}: 
\begin{align}
    G-Mean = \sqrt[2]{\frac{TP}{TP+FN} \times \frac{TN}{TN+FP}} 
\end{align}
where $\frac{TP}{TP+FN}$ is the positive class recall and $\frac{TN}{TN+FP}$ is the negative class recall. Macro and micro G-Mean are obtained by using G-Mean as the performance metric in Equations \ref{eq: marco} and \ref{eq: micro}. Unlike recall, precision, or F-Score, G-Mean offers a more balanced assessment of model performance across both classes for each label.

Label-set-based G-Mean (LS-G-Mean) is computed by calculating $TP$, $FP$, $TN$, and $FN$ on each received label set:
\begin{gather}
    \nonumber G-Mean_{LS} = \\ \frac{1}{|D|}\sum_{t=1}^{|D|}G-Mean(TP^{t}, FP^{t}, TN^{t}, FN^{t})
\end{gather}
where $TP^{t}$, $FP^{t}$, $TN^{t}$, and $FN^{t}$ are the confusion matrix counts computed for the predicted $\hat{\mathbf{y}}^t$ and true label sets $\mathbf{y}^t$. Compared to HScore and HLoss, LS-G-Mean gives equal weight to all classes across label sets, helping to avoid the effects of class imbalance. 

\section{Experimental Results and Analysis}
\label{sec: Experimental Results and Analysis}
\begin{table*}[tb]\centering
\caption{Experimental results on each dataset. BR-M refers to BR-MARLENE, BRPW-M refers to BRPW-MARLENE, Macro refers to Macro-G-Mean, Micro to Micro-G-Mean, LS to Label-set-based-G-Mean, and Rank to Friedman’s rank. Friedman’s p-values are always $<2.2 \times 10^{-16}$. The best approach is highlighted in red, and approaches without a significant difference according to the post-hoc Nemenyi test are shown in bold.}
\scalebox{0.85}{
\begin{tabular}{|l|l|l|l|l|l|l|l|l|l|l|l|l|l|l|l|}
\hline
Dataset   & \multicolumn{2}{l|}{Slashdot} & \multicolumn{2}{l|}{Ohsumed} & \multicolumn{2}{l|}{Reuters} & \multicolumn{2}{l|}{Yeast} & \multicolumn{2}{l|}{20NG} & \multicolumn{2}{l|}{TMC2007} & \multicolumn{2}{l|}{IMDB} &           \\ \hlineB{5}
Metrics   & Macro         & Rank          & Macro        & Rank          & Macro        & Rank          & Macro       & Rank         & Macro       & Rank        & Macro        & Rank          & Macro       & Rank        & Avg. Rank \\ \hline
\textbf{BRPW-M} & - & - & - & - & - & - & 0.567 & -& - & - & - & - & - & - & - \\ 
\textbf{BR-M} & \cellcolor[gray]{0.9}\textbf{\textcolor{red}{0.357}}         & \cellcolor[gray]{0.9}\textbf{\textcolor{red}{1.116}}        & \cellcolor[gray]{0.9}\textbf{\textcolor{red}{0.571}}        & \cellcolor[gray]{0.9}\textbf{\textcolor{red}{1.019}}         & \cellcolor[gray]{0.9}\textbf{\textcolor{red}{0.415}}       & \cellcolor[gray]{0.9}\textbf{\textcolor{red}{1.003}}         & \cellcolor[gray]{0.9}\textbf{\textcolor{red}{0.539}}       & \cellcolor[gray]{0.9}\textbf{\textcolor{red}{1.012}}        & \cellcolor[gray]{0.9}\textbf{\textcolor{red}{0.699}}       & \cellcolor[gray]{0.9}\textbf{\textcolor{red}{1.023}}       & \cellcolor[gray]{0.9}\textbf{\textcolor{red}{0.752}}        & \cellcolor[gray]{0.9}\textbf{\textcolor{red}{1.003}}         & \cellcolor[gray]{0.9}\textbf{\textcolor{red}{0.411}}       & \cellcolor[gray]{0.9}\textbf{\textcolor{red}{1.000}}       & \cellcolor[gray]{0.9}\textbf{\textcolor{red}{1.025}}     \\ 
EBR       & 0.023         & 7.850         & 0.294        & 3.662         & 0.048        & 5.877         & 0.232       & 6.464        & 0.506       & 3.075       & 0.491        & 2.203         & 0.059       & 4.271       & 4.772     \\ 
ECC       & 0.020         & 8.992         & 0.273        & 4.828         & 0.041        & 6.757         & 0.239       & 5.802        & 0.490       & 4.587       & 0.479        & 3.172         & 0.035       & 6.441       & 5.797     \\ 
EPS       & 0.081         & 6.356         & 0.151        & 6.073         & 0.035        & 6.642         & 0.295       & 2.542        & 0.268       & 7.138       & 0.245        & 6.443         & 0.008       & 9.412       & 6.372     \\ 
ERT      & 0.000         & 10.290        & 0.030        & 10.477        & 0.000        & 10.718        & 0.000       & 11.991       & 0.112       & 11.101      & 0.006        & 12.000        & 0.002       & 11.273      & 11.121    \\ 
EaBR      & 0.010         & 8.776         & 0.227        & 5.893         & 0.020        & 8.501         & 0.231       & 6.698        & 0.464       & 5.095       & 0.436        & 3.979         & 0.031       & 6.789       & 6.533     \\ 
EaCC      & 0.019         & 9.313         & 0.016        & 11.278        & 0.003        & 10.123        & 0.238       & 6.077        & 0.270       & 7.713       & 0.411        & 4.946         & 0.006       & 10.442      & 8.556     \\ 
EaPS      & 0.110         & 3.996         & 0.126        & 7.327         & 0.049        & 4.812         & 0.288       & 3.598        & 0.199       & 9.364       & 0.246        & 6.345         & 0.012       & 9.424       & 6.409     \\ 
GOBR      & 0.027         & 7.143         & 0.126        & 7.293         & 0.042        & 5.700         & 0.221       & 7.178        & 0.294       & 6.355       & 0.177        & 7.979         & 0.126       & 2.751       & 6.343     \\ 
GOCC      & 0.028         & 7.106         & 0.134        & 6.949         & 0.013        & 9.327         & 0.196       & 9.046        & 0.176       & 9.779       & 0.163        & 9.052         & 0.154       & 2.249       & 7.644     \\ 
GOPS      & 0.196         & 2.415         & 0.196        & 4.558         & 0.047        & 4.704         & 0.218       & 7.358        & 0.271       & 6.523       & 0.138        & 9.878         & 0.014       & 8.685       & 6.303     \\ 
GORT      & 0.068         & 4.647         & 0.092        & 8.642         & 0.060        & 3.837         & 0.177       & 10.233       & 0.293       & 6.247       & 0.090        & 11.000        & 0.043       & 5.263       & 7.124     \\ \hlineB{5}
Metrics   & Micro & Rank   & Micro & Rank   & Micro & Rank   & Micro & Rank   & Micro & Rank   & Micro & Rank   & Micro & Rank   & Avg. Rank \\ \hline
\textbf{BRPW-M} & - & - & - & - & - & - & 0.663 & -& - & - & - & - & - & - & - \\ 
\textbf{BR-M} & 0.572 & 2.362  & \cellcolor[gray]{0.9}\textbf{\textcolor{red}{0.610}} & \cellcolor[gray]{0.9}\textbf{\textcolor{red}{1.461}}  & \cellcolor[gray]{0.9}\textbf{\textcolor{red}{0.753}} & \cellcolor[gray]{0.9}\textbf{\textcolor{red}{1.000}}  & 0.644 & 8.694  & \cellcolor[gray]{0.9}\textbf{\textcolor{red}{0.759}} & \cellcolor[gray]{0.9}\textbf{\textcolor{red}{1.004}}  & \cellcolor[gray]{0.9}\textbf{\textcolor{red}{0.774}} & 
\cellcolor[gray]{0.9}\textbf{\textcolor{red}{1.763}}  & 0.524 & 2.611  & \cellcolor[gray]{0.9}\textbf{\textcolor{red}{2.699}}     \\ 
EBR       & 0.065 & 8.692  & 0.375 & 5.342  & 0.188 & 6.292  & \cellcolor[gray]{0.9}\textbf{\textcolor{red}{0.724}} & \cellcolor[gray]{0.9}\textbf{\textcolor{red}{2.165}}  & 0.561 & 3.131  & 0.761 & 2.379  & 0.226 & 8.034  & 5.148     \\ 
ECC       & 0.059 & 9.715  & 0.360 & 6.524  & 0.177 & 7.325  & 0.718 & 2.593  & 0.547 & 4.652  & 0.754 & 3.530  & 0.111 & 9.667  & 6.287     \\ 
EPS       & 0.333 & 5.076  & 0.458 & 5.410  & 0.335 & 5.654  & 0.697 & 5.883  & 0.462 & 5.835  & 0.691 & 6.851  & 0.414 & 5.501  & 5.744     \\ 
ERT      & 0.003 & 10.817 & 0.165 & 10.495 & 0.001 & 10.813 & 0.000 & 12.000 & 0.255 & 10.094 & 0.039 & 12.000 & 0.013 & 11.565 & 11.112    \\ 
EaBR      & 0.040 & 9.572  & 0.309 & 7.727  & 0.112 & 9.198  & \cellcolor[gray]{0.9}\textbf{\textcolor{red}{0.724}} & \cellcolor[gray]{0.9}\textbf{\textcolor{red}{2.445}}  & 0.523 & 5.148  & 0.755 & 2.784  & 0.133 & 9.327  & 6.600     \\ 
EaCC      & 0.057 & 9.987  & 0.037 & 11.626 & 0.024 & 10.199 & 0.717 & 2.951  & 0.337 & 8.172  & 0.745 & 4.571  & 0.021 & 11.407 & 8.416     \\ 
EaPS      & 0.287 & 5.975  & 0.415 & 7.070  & 0.380 & 3.882  & 0.700 & 5.015  & 0.358 & 8.655  & 0.682 & 6.480  & 0.415 & 5.837  & 6.131     \\ 
GOBR      & \cellcolor[gray]{0.9}\textbf{\textcolor{red}{0.598}} & \cellcolor[gray]{0.9}\textbf{\textcolor{red}{1.877}}  & 0.546 & 2.933  & 0.407 & 4.309  & 0.576 & 9.246  & 0.448 & 6.239  & 0.554 & 8.924  & 0.570 & 2.154  & 5.097     \\ 
GOCC      & 0.155 & 7.333  & 0.256 & 9.165  & 0.081 & 9.448  & 0.577 & 9.645  & 0.230 & 10.871 & 0.531 & 10.439 & 0.372 & 5.683  & 8.941     \\ 
GOPS      & 0.365 & 4.564  & 0.414 & 6.567  & 0.282 & 6.715  & 0.576 & 9.468  & 0.337 & 8.578  & 0.521 & 10.535 & 0.429 & 4.921  & 7.335     \\ 
GORT      & \cellcolor[gray]{0.9}\textbf{0.596} & \cellcolor[gray]{0.9}\textbf{2.031}  & 0.506 & 3.680  & 0.408 & 3.164  & 0.633 & 7.895  & 0.487 & 5.620  & 0.640 & 7.745  & \cellcolor[gray]{0.9}\textbf{\textcolor{red}{0.612}} & \cellcolor[gray]{0.9}\textbf{\textcolor{red}{1.295}}  & 4.490     \\ \hlineB{5}
Metrics   & LS    & Rank   & LS    & Rank   & LS    & Rank   & LS    & Rank   & LS    & Rank   & LS    & Rank   & LS    & Rank   & Avg. Rank \\ \hline
\textbf{BRPW-M} & - & - & - & - & - & - & 0.636 & -& - & - & - & - & - & - & - \\ 
\textbf{BR-M} & 0.412 & 2.345  & \cellcolor[gray]{0.9}\textbf{\textcolor{red}{0.497}} & \cellcolor[gray]{0.9}\textbf{\textcolor{red}{1.347}}  & \cellcolor[gray]{0.9}\textbf{\textcolor{red}{0.640}} & \cellcolor[gray]{0.9}\textbf{\textcolor{red}{1.000}}  & 0.606 & 8.404  & \cellcolor[gray]{0.9}\textbf{\textcolor{red}{0.650}} & \cellcolor[gray]{0.9}\textbf{\textcolor{red}{1.012}}  & \cellcolor[gray]{0.9}\textbf{\textcolor{red}{0.733}} & \cellcolor[gray]{0.9}\textbf{\textcolor{red}{2.120}}  & 0.397 & 2.615  & \cellcolor[gray]{0.9}\textbf{\textcolor{red}{2.692}}     \\ 
EBR       & 0.015 & 8.559  & 0.234 & 5.321  & 0.097 & 6.290  & \cellcolor[gray]{0.9}\textbf{\textcolor{red}{0.697}} & \cellcolor[gray]{0.9}\textbf{\textcolor{red}{1.912}}  & 0.363 & 3.157  & 0.720 & 2.231  & 0.079 & 8.029  & 5.071     \\ 
ECC       & 0.013 & 9.667  & 0.221 & 6.509  & 0.089 & 7.188  & 0.685 & 2.902  & 0.347 & 4.732  & 0.707 & 3.835  & 0.018 & 9.765  & 6.371     \\ 
EPS       & 0.134 & 5.050  & 0.275 & 5.242  & 0.157 & 5.431  & 0.646 & 6.045  & 0.223 & 5.859  & 0.630 & 6.925  & 0.261 & 5.361  & 5.702     \\ 
ERT      & 0.000 & 10.828 & 0.050 & 10.461 & 0.000 & 10.813 & 0.000 & 12.000 & 0.082 & 10.120 & 0.004 & 12.000 & 0.000 & 11.600 & 11.117    \\ 
EaBR      & 0.006 & 9.832  & 0.178 & 7.743  & 0.046 & 9.116  & 0.697 & 2.374  & 0.332 & 5.075  & 0.717 & 2.516  & 0.030 & 9.233  & 6.556     \\ 
EaCC      & 0.012 & 9.942  & 0.005 & 11.629 & 0.004 & 10.170 & 0.684 & 2.922  & 0.152 & 8.182  & 0.700 & 4.559  & 0.001 & 11.372 & 8.396     \\ 
EaPS      & 0.101 & 5.969  & 0.230 & 6.953  & 0.205 & 3.448  & 0.652 & 5.169  & 0.138 & 8.760  & 0.619 & 6.495  & 0.262 & 5.707  & 6.071     \\ 
GOBR      & \cellcolor[gray]{0.9}\textbf{\textcolor{red}{0.436}} & \cellcolor[gray]{0.9}\textbf{\textcolor{red}{1.923}}  & 0.381 & 3.084  & 0.216 & 4.729  & 0.517 & 8.830  & 0.224 & 6.285  & 0.436 & 8.910  & 0.451 & 2.235  & 5.142     \\ 
GOCC      & 0.031 & 7.336  & 0.087 & 9.177  & 0.015 & 9.557  & 0.517 & 9.658  & 0.057 & 10.887 & 0.405 & 10.416 & 0.208 & 6.020  & 9.007     \\ 
GOPS      & 0.156 & 4.565  & 0.223 & 6.750  & 0.112 & 6.738  & 0.514 & 10.046 & 0.126 & 8.367  & 0.392 & 10.549 & 0.284 & 4.740  & 7.393     \\ 
GORT      & \cellcolor[gray]{0.9}\textbf{0.426} & \cellcolor[gray]{0.9}\textbf{1.984}  & 0.321 & 3.785  & 0.206 & 3.521  & 0.594 & 7.739  & 0.267 & 5.565  & 0.570 & 7.445  & \cellcolor[gray]{0.9}\textbf{\textcolor{red}{0.515}} & \cellcolor[gray]{0.9}\textbf{\textcolor{red}{1.322}}  & 4.480     \\ \hline
\end{tabular}}
\label{table: performance results table}
\end{table*}

\subsection{Results on Real-World Datasets}
\label{sec: Results on Real-World Datasets}
The predictive performance of each algorithm on each dataset is shown in Table \ref{table: performance results table}. Due to the high computational cost from modelling pairwise dependencies, BRPW-MARLENE is only evaluated on the Yeast dataset ($|\mathcal{L}|=14$). As BRPW-MARLENE is an extension of BR-MARLENE and only tested on Yeast, we first compare BR-MARLENE with other benchmarks, and then compare BRPW-MARLENE with BR-MARLENE.

BR-MARLENE achieves the best average Friedman rank and significantly higher Macro-G-Mean values than all other approaches across datasets. Since Macro-G-Mean is particularly sensitive to minority class results of each label (as discussed in Section \ref{sec: Evaluation Metrics}) and all datasets have very low $LIR$ (see Table \ref{table: real-world datasets}), this demonstrates that BR-MARLENE is especially effective at handling class imbalance for each label.

BR-MARLENE achieves the best average Friedman ranking in Micro-G-Mean across all datasets, reflecting strong global performance over all classes and labels. While it does not have the top Micro-G-Mean on three datasets (Slashdot, Yeast, and IMDB) because it assigns less weight to the majority class, it still delivers competitive results (e.g., BR-MARLENE’s Micro-G-Mean on Slashdot is 0.572, compared to 0.598 for the best method GOBR). Results for LS-G-Mean are similar: BR-MARLENE has the best average Friedman ranking and the highest ranking position on most datasets.

The average runtime for each approach is shown in Table \ref{table: execution time}. BR-MARLENE has the shortest execution time on all datasets except Reuters, TMC2007, and IMDB, as these datasets have more labels and may contain more concept drift, increasing the number of sub-classifiers. Nevertheless, BR-MARLENE remains the second fastest on TMC2007 and third fastest on Reuters and IMDB.

Comparing BR-MARLENE and BRPW-MARLENE on Yeast, BRPW-MARLENE achieves higher performance across all three evaluation metrics, demonstrating that modelling label dependencies can indeed improve performance. However, BRPW-MARLENE is nearly 53 times slower (see Table \ref{table: execution time}) than BR-MARLENE due to the large number of pairwise models. More comparisons between BR-MARLENE and BRPW-MARLENE on synthetic datasets are discussed in the next section.

\begin{table}[tb]\centering
\caption{Tabulations of execution time. All values are in milliseconds. The best values of each column are in red.}
\begin{center}
\scalebox{0.75}{

\begin{tabular}{|l|l|l|l|l|l|l|l|}
\hline
Dataset   & Slashdot & Ohsumed & Reuters & Yeast & 20NG    & TMC2007 & IMDB     \\ \hline
\textbf{BRPW-M} & -        & -       & -       & 82604 & -     & -       & -   \\
\textbf{BR-M} & \cellcolor[gray]{0.9}\textbf{\textcolor{red}{24111}}    & \cellcolor[gray]{0.9}\textbf{\textcolor{red}{110207}}  & 88068   & \cellcolor[gray]{0.9}\textbf{\textcolor{red}{1562}}  & \cellcolor[gray]{0.9}\textbf{\textcolor{red}{134837}}  & 105846  & 5754256  \\ 
EBR       & 400567   & 1932007 & 1958737 & 20823 & 2284106 & 2000631 & 31679670 \\ 
ECC       & 400471   & 1804186 & 1947755 & 20072 & 2269085 & 1994518 & 31296078 \\ 
EPS       & 64995    & 185806  & \cellcolor[gray]{0.9}\textbf{\textcolor{red}{34768}}   & 3961  & 273480  & \cellcolor[gray]{0.9}\textbf{\textcolor{red}{78551}}   & \cellcolor[gray]{0.9}\textbf{\textcolor{red}{296169}}   \\ 
ERT      & 53813    & 238408  & 186590  & 8090  & 259720  & 225389  & 1895433  \\ 
EaBR      & 394570   & 2049120 & 2100270 & 32151 & 2560537 & 2427817 & 22042432 \\ 
EaCC      & 391400   & 1757911 & 2011307 & 32178 & 2214708 & 2337822 & 18848438 \\ 
EaPS      & 169128   & 424801  & 112571  & 8383  & 765386  & 243306  & 639729   \\ 
GOBR      & 559366   & 2001526 & 2024646 & 22191 & 2640190 & 2179204 & 26444018 \\ 
GOCC      & 565521   & 2015383 & 2221846 & 25066 & 2643139 & 2281488 & 27882590 \\ 
GOPS      & 113790   & 269166  & 76479   & 7536  & 432014  & 141485  & 546422   \\ 
GORT      & 75365    & 294842  & 209673  & 7824  & 367842  & 334401  & 3200813  \\ \hline
\end{tabular}
}\end{center}
\label{table: execution time}
\end{table}

\subsection{Results on Synthetic Datasets}
\label{sec: Results on Synthetic Datasets}
To further investigate the impact of multiple sources and different concept drift types, we run BR/BRPW-MARLENE on several synthetic datasets. The Friedman ranks for each evaluation metric are reported in \ref{table: friedman on synthetic dataset}.
BRPW-MARLENE with similar sources, non-similar sources, and no source achieve the first, second, and third best average ranks across all metrics, respectively. This shows that modelling label dependencies with BRPW-MARLENE consistently outperforms all BR-MARLENE with/without sources. Moreover, BRPW-MARLENE benefits most from similar sources, but non-similar sources can also be helpful.

Among BR-MARLENE methods using different source settings, incorporating similar sources gives the best average performance across all metrics, with no-source BR-MARLENE second best on Macro-G-Mean and Micro-G-Mean, and non-similar sources second on LS-G-Mean. This shows that similar sources offer greater improvement to BR-MARLENE, as also observed for BRPW-MARLENE. It is worth noting that BR/BRPW-MARLENE with similar sources do not always achieve the best ranks (e.g., Macro AA 5000); as the number of target examples increases, both methods can perform well even without source help. Thus, similar sources are most beneficial when the target size is small, which is further investigated in the next section.

\begin{table}[tb]
\caption{Friedman’s ranks on synthetic datasets: Macro, Micro, and LS refer to Macro-G-Mean, Micro-G-Mean, and Label-set-based-G-Mean, respectively. Best ranks are in red, bold indicates no significant difference by Nemenyi test, and all Friedman’s p-values are always $<2.2 \times 10^{-16}$}
\begin{center}
\scalebox{0.65}{
\begin{tabular}{|l|l|l l l l l l|}
\hline
\multicolumn{2}{|l|}{Dataset}             & BR w/o S & BRPW w/o S & BR w/ NS & BRPW w/ NS & BR w/ SS & BRPW w/ SS \\ \hlineB{5}
Drift Type             & Size   & \multicolumn{6}{l|}{Macro}                                                                \\ \hline
\multirow{3}{*}{SS}    & 50     & 4.392     & \cellcolor[gray]{0.9}\textbf{2.768}       & 4.292         & \cellcolor[gray]{0.9}\textbf{2.926}           & 4.152        & \cellcolor[gray]{0.9}\textbf{\textcolor{red}{2.470}}          \\  
                      & 500    & 5.315     & 2.501       & 5.114         & 2.981           & 3.308        & \cellcolor[gray]{0.9}\textbf{\textcolor{red}{1.781}}          \\  
                      & 5000   & 5.038     & 1.858       & 5.752         & 2.692           & 4.092        & \cellcolor[gray]{0.9}\textbf{\textcolor{red}{1.568}}          \\ \hline
\multirow{3}{*}{IS}    & 50     & 4.522     & 2.966       & 4.853         & 2.748           & 3.796        & \cellcolor[gray]{0.9}\textbf{\textcolor{red}{2.115}}          \\  
                      & 500    & 5.155     & 2.345       & 5.616         & 2.716           & 3.664        & \cellcolor[gray]{0.9}\textbf{\textcolor{red}{1.503}}          \\  
                      & 5000   & 4.714     & \cellcolor[gray]{0.9}\textbf{\textcolor{red}{1.765}}       & 5.444         & 2.581           & 4.470        & 2.025          \\ \hline
\multirow{3}{*}{II}    & 50     & 5.212     & 2.808       & 5.171         & 2.245           & 3.986        & \cellcolor[gray]{0.9}\textbf{\textcolor{red}{1.578}}          \\  
                      & 500    & 5.081     & 3.094       & 5.323         & 2.189           & 3.802        & \cellcolor[gray]{0.9}\textbf{\textcolor{red}{1.512}}          \\  
                      & 5000   & 4.853     & 2.847       & 5.214         & 1.893           & 4.579        & \cellcolor[gray]{0.9}\textbf{\textcolor{red}{1.614}}          \\ \hline
\multirow{3}{*}{IA}    & 50     & 5.038     & 2.965       & 5.133         & 2.884           & 3.682        & \cellcolor[gray]{0.9}\textbf{\textcolor{red}{1.298}}          \\  
                      & 500    & 4.877     & 2.619       & 5.216         & 2.461           & 4.235        & \cellcolor[gray]{0.9}\textbf{\textcolor{red}{1.591}}          \\  
                      & 5000   & 4.641     & 2.501       & 5.248         & 1.975           & 4.852        & \cellcolor[gray]{0.9}\textbf{\textcolor{red}{1.783}}          \\ \hline
\multirow{3}{*}{AA}    & 50     & 4.712     & 2.977       & 4.539         & \cellcolor[gray]{0.9}\textbf{\textcolor{red}{2.456}}           & 3.779        & \cellcolor[gray]{0.9}\textbf{2.537}          \\  
                      & 500    & 5.104     & 2.889       & 4.817         & \cellcolor[gray]{0.9}\textbf{\textcolor{red}{1.613}}           & 4.514        & 2.064          \\  
                      & 5000   & 4.996     & \cellcolor[gray]{0.9}\textbf{\textcolor{red}{1.793}}       & 5.291         & 2.233           & 4.525        & 2.162          \\ \hline
\multirow{3}{*}{AS}    & 50     & 4.228     & 3.255       & 4.374         & 3.200           & 3.507        & \cellcolor[gray]{0.9}\textbf{\textcolor{red}{2.436}}          \\  
                      & 500    & 5.084     & \cellcolor[gray]{0.9}\textbf{2.380}       & 4.994         & 3.070           & 3.172        & \cellcolor[gray]{0.9}\textbf{\textcolor{red}{2.300}}          \\  
                      & 5000   & 4.955     & 1.957       & 5.860         & \cellcolor[gray]{0.9}\textbf{\textcolor{red}{1.691}}           & 4.038        & 2.499          \\ \hline
\multicolumn{2}{|l|}{Avg. Rank} & 4.884     & 2.571       & 5.125         & 2.475           & 4.008        & \cellcolor[gray]{0.9}\textbf{\textcolor{red}{1.935}}          \\ \hlineB{5}
Drift Type             & Size   & \multicolumn{6}{l|}{Micro}                                                                \\ \hline
\multirow{3}{*}{SS}    & 50     & 4.380     & \cellcolor[gray]{0.9}\textbf{2.804 }      & 4.280         & \cellcolor[gray]{0.9}\textbf{2.910}           & 4.094        & \cellcolor[gray]{0.9}\textbf{\textcolor{red}{2.532}}          \\  
                      & 500    & 5.363     & 2.846       & 5.091         & 2.579           & 3.193        & \cellcolor[gray]{0.9}\textbf{\textcolor{red}{1.928}}          \\  
                      & 5000   & 4.891     & 2.376       & 5.698         & \cellcolor[gray]{0.9}\textbf{\textcolor{red}{1.724}}           & 4.298        & 2.013          \\ \hline
\multirow{3}{*}{IS}    & 50     & 4.226     & 3.244       & 4.538         & 3.032           & 3.704        & \cellcolor[gray]{0.9}\textbf{\textcolor{red}{2.256}}          \\  
                      & 500    & 5.065     & 2.648       & 5.533         & 2.757           & 3.521        & \cellcolor[gray]{0.9}\textbf{\textcolor{red}{1.477}}          \\  
                      & 5000   & 4.767     & \cellcolor[gray]{0.9}\textbf{1.986}       & 5.469         & 2.353           & 4.447        & \cellcolor[gray]{0.9}\textbf{\textcolor{red}{1.978}}          \\ \hline
\multirow{3}{*}{II}    & 50     & 4.996     & 2.979       & 4.986         & 2.407           & 3.964        & \cellcolor[gray]{0.9}\textbf{\textcolor{red}{1.668}}          \\  
                      & 500    & 4.706     & 3.239       & 5.087         & 2.270           & 4.252        & \cellcolor[gray]{0.9}\textbf{\textcolor{red}{1.445}}          \\  
                      & 5000   & 4.802     & 2.953       & 5.090         & 1.810           & 4.766        & \cellcolor[gray]{0.9}\textbf{\textcolor{red}{1.578}}          \\ \hline
\multirow{3}{*}{IA}    & 50     & 4.943     & 3.002       & 5.189         & 2.988           & 3.591        & \cellcolor[gray]{0.9}\textbf{\textcolor{red}{1.288}}          \\  
                      & 500    & 4.651     & 2.806       & 5.058         & 2.807           & 4.149        & \cellcolor[gray]{0.9}\textbf{\textcolor{red}{1.529}}          \\  
                      & 5000   & 4.561     & 2.640       & 5.222         & 1.896           & 4.955        & \cellcolor[gray]{0.9}\textbf{\textcolor{red}{1.726}}          \\ \hline
\multirow{3}{*}{AA}    & 50     & 4.689     & 3.015       & 4.510         & \cellcolor[gray]{0.9}\textbf{\textcolor{red}{2.348}}           & 3.692        & 2.746          \\  
                      & 500    & 5.005     & 3.326       & 4.678         & \cellcolor[gray]{0.9}\textbf{\textcolor{red}{1.499}}           & 4.250        & 2.241          \\  
                      & 5000   & 4.907     & 2.410       & 5.045         & \cellcolor[gray]{0.9}\textbf{\textcolor{red}{1.404}}           & 4.867        & 2.367          \\ \hline
\multirow{3}{*}{AS}    & 50     & 4.078     & 3.533       & 4.290         & 3.251           & 3.242        & \cellcolor[gray]{0.9}\textbf{\textcolor{red}{2.606}}          \\  
                      & 500    & 5.051     & \cellcolor[gray]{0.9}\textbf{\textcolor{red}{2.535}}       & 4.880         & 2.664           & 3.304        & \cellcolor[gray]{0.9}\textbf{2.566}          \\  
                      & 5000   & 4.947     & 2.296       & 5.867         & \cellcolor[gray]{0.9}\textbf{\textcolor{red}{1.355}}           & 3.996        & 2.538          \\ \hline
\multicolumn{2}{|l|}{Avg. Rank} & 4.779     & 2.813       & 5.028         & 2.336           & 4.016        & \cellcolor[gray]{0.9}\textbf{\textcolor{red}{2.027}}          \\ \hlineB{5}
Drift Type             & Size   & \multicolumn{6}{l|}{LS}                                                                \\ \hline
\multirow{3}{*}{SS}    & 50     & 4.528     & \cellcolor[gray]{0.9}\textbf{2.890}       & 4.428         & \cellcolor[gray]{0.9}\textbf{2.856}           & 3.864        & \cellcolor[gray]{0.9}\textbf{\textcolor{red}{2.434}}          \\  
                      & 500    & 5.406     & 3.276       & 4.924         & \cellcolor[gray]{0.9}\textbf{2.090}           & 3.308        & \cellcolor[gray]{0.9}\textbf{\textcolor{red}{1.995}}          \\  
                      & 5000   & 5.344     & 2.767       & 5.140         & \cellcolor[gray]{0.9}\textbf{\textcolor{red}{1.164}}           & 4.353        & 2.233          \\ \hline
\multirow{3}{*}{IS}    & 50     & 4.154     & 3.509       & 4.333         & 3.033           & 3.330        & \cellcolor[gray]{0.9}\textbf{\textcolor{red}{2.642}}          \\  
                      & 500    & 5.416     & 2.965       & 5.180         & 2.761           & 3.451        & \cellcolor[gray]{0.9}\textbf{\textcolor{red}{1.227}}          \\  
                      & 5000   & 5.097     & 2.234       & 5.297         & \cellcolor[gray]{0.9}\textbf{\textcolor{red}{1.920}}           & 4.461        & 1.990          \\ \hline
\multirow{3}{*}{II}    & 50     & 4.538     & 3.654       & 4.581         & 2.964           & 3.136        & \cellcolor[gray]{0.9}\textbf{\textcolor{red}{2.128}}          \\  
                      & 500    & 4.912     & 3.078       & 5.184         & 2.259           & 4.320        & \cellcolor[gray]{0.9}\textbf{\textcolor{red}{1.247}}          \\  
                      & 5000   & 5.360     & 3.031       & 4.539         & 1.692           & 4.955        & \cellcolor[gray]{0.9}\textbf{\textcolor{red}{1.424}}          \\ \hline
\multirow{3}{*}{IA}    & 50     & 4.757     & 3.360       & 4.927         & 3.207           & 3.262        & \cellcolor[gray]{0.9}\textbf{\textcolor{red}{1.486}}          \\  
                      & 500    & 5.118     & 2.850       & 4.793         & 3.142           & 3.724        & \cellcolor[gray]{0.9}\textbf{\textcolor{red}{1.372}}          \\  
                      & 5000   & 5.280     & 2.378       & 4.760         & 2.179           & 4.795        & \cellcolor[gray]{0.9}\textbf{\textcolor{red}{1.608}}          \\ \hline
\multirow{3}{*}{AA}    & 50     & 4.436     & 3.288       & 4.190         & \cellcolor[gray]{0.9}\textbf{\textcolor{red}{2.386}}           & 3.663        & 3.037          \\  
                      & 500    & 5.376     & 3.632       & 5.051         & \cellcolor[gray]{0.9}\textbf{\textcolor{red}{1.628}}           & 3.201        & 2.113          \\  
                      & 5000   & 5.151     & 2.476       & 4.364         & \cellcolor[gray]{0.9}\textbf{\textcolor{red}{1.035}}           & 5.301        & 2.673          \\ \hline
\multirow{3}{*}{AS}    & 50     & 4.143     & 3.809       & 4.195         & \cellcolor[gray]{0.9}\textbf{3.134}           & \cellcolor[gray]{0.9}\textbf{2.917}        & \cellcolor[gray]{0.9}\textbf{\textcolor{red}{2.802}}          \\  
                      & 500    & 5.172     & 3.088       & 4.867         & \cellcolor[gray]{0.9}\textbf{\textcolor{red}{2.098}}           & 3.487        & 2.288          \\  
                      & 5000   & 5.646     & 2.736       & 5.008         & \cellcolor[gray]{0.9}\textbf{\textcolor{red}{1.139}}           & 4.037        & 2.433          \\ \hline
\multicolumn{2}{|l|}{Avg. Rank} & 4.991     & 3.057       & 4.764         & 2.260           & 3.865        & \cellcolor[gray]{0.9}\textbf{\textcolor{red}{2.063}}          \\ \hline
\end{tabular}}\end{center}
\label{table: friedman on synthetic dataset}
\end{table}

\begin{figure*}[tb]
\centering
\begin{subfigure}{0.235\textwidth}
    \includegraphics[scale=0.14]{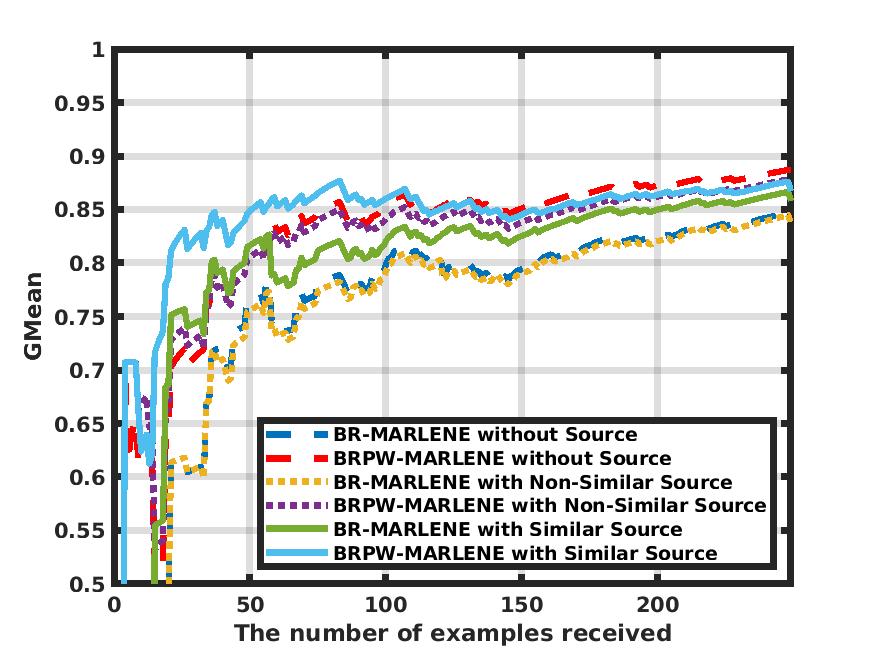}
    \caption{SS; $l_{T,1}$; size of 50}
    \label{subfig:ss_1}
\end{subfigure}
\begin{subfigure}{0.235\textwidth}
    \includegraphics[scale=0.14]{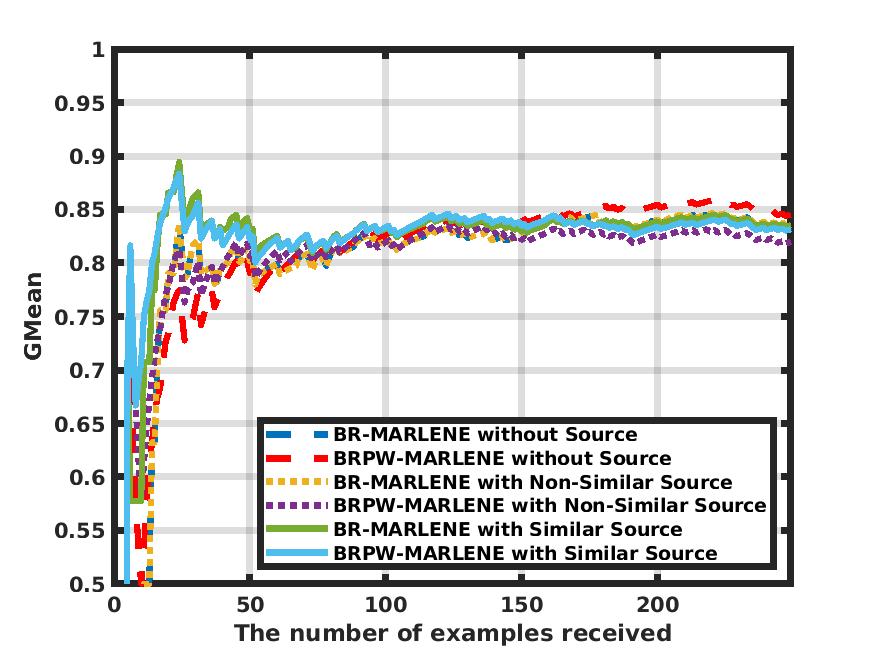}
    \caption{SS; $l_{T,2}$; size of 50}
    \label{subfig:ss_2}
\end{subfigure}
\begin{subfigure}{0.235\textwidth}
    \includegraphics[scale=0.14]{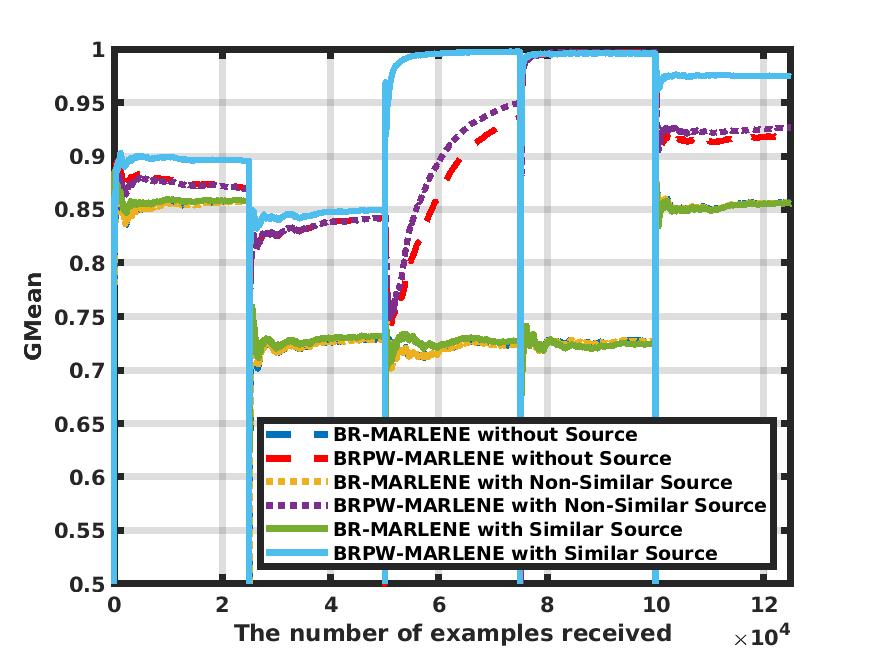}
    \caption{IS; $l_{T,1}$; size of 5000}
    \label{subfig:is_1}
\end{subfigure}
\begin{subfigure}{0.235\textwidth}
    \includegraphics[scale=0.14]{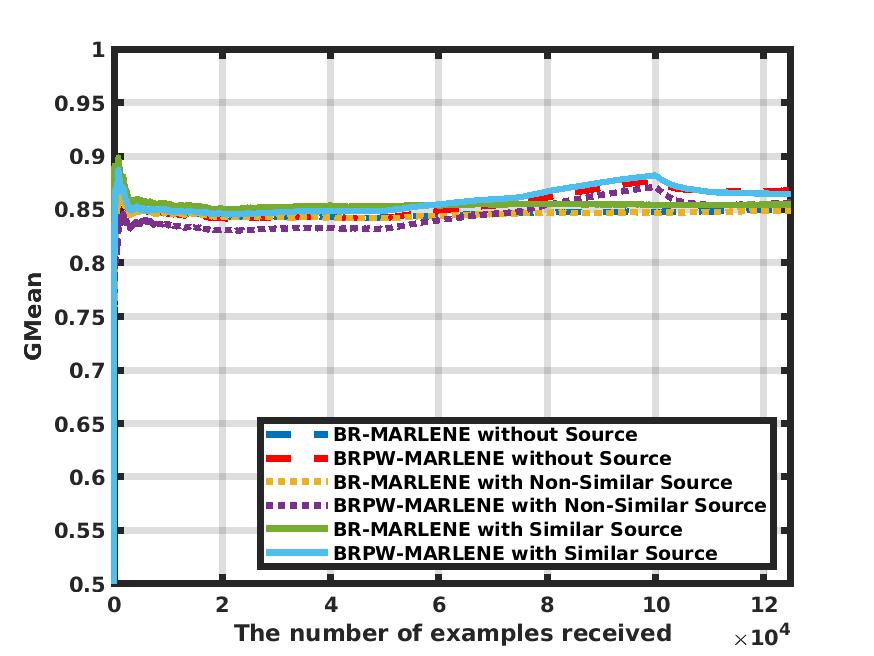}
    \caption{IS; $l_{T,2}$; size of 5000}
    \label{subfig:is_2}
\end{subfigure}

\vspace{1ex}

\begin{subfigure}{0.235\textwidth}
    \includegraphics[scale=0.14]{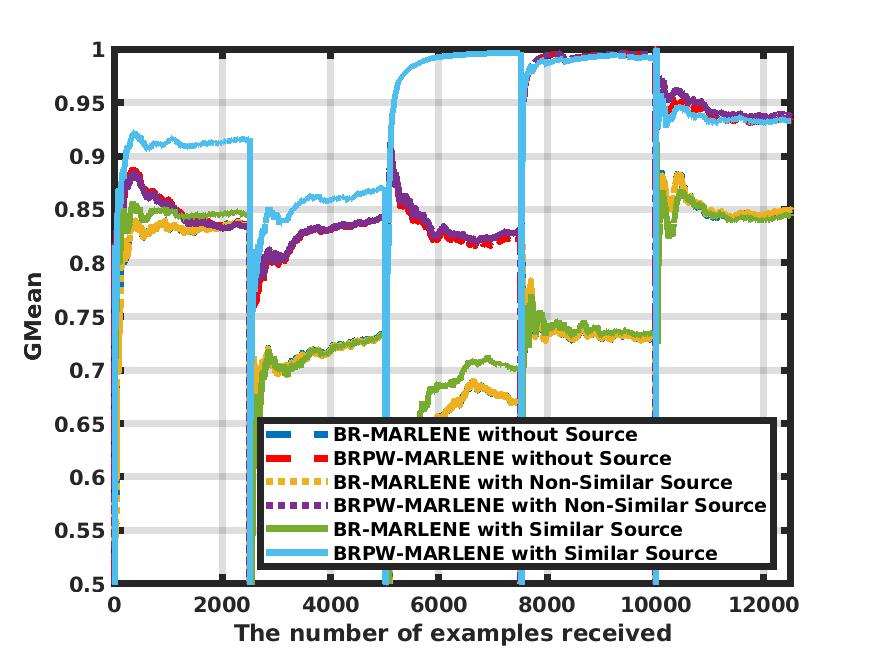}
    \caption{II; $l_{T,1}$; size of 500}
    \label{subfig:ii_1}
\end{subfigure}
\begin{subfigure}{0.235\textwidth}
    \includegraphics[scale=0.14]{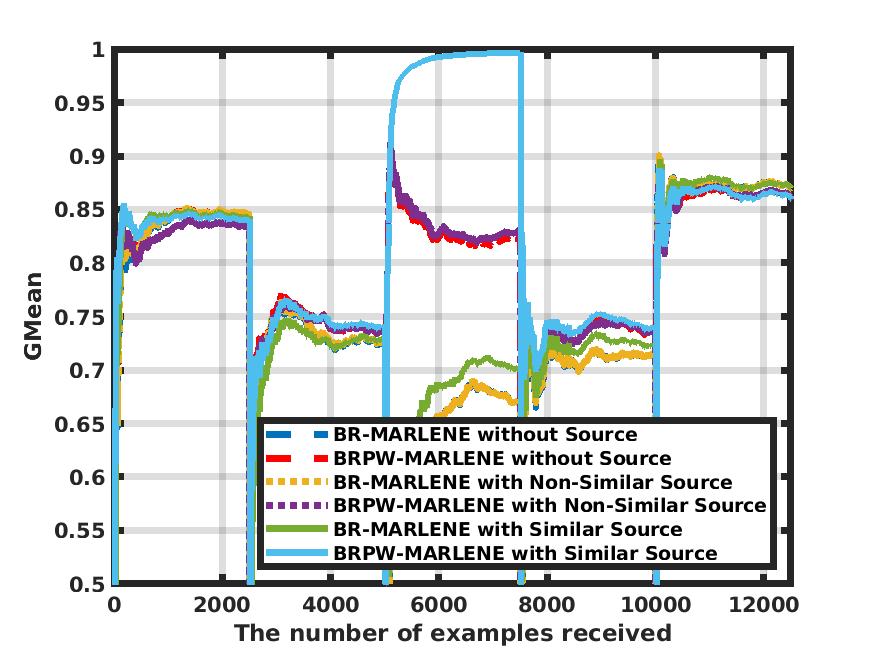}
    \caption{II; $l_{T,2}$; size of 500}
    \label{subfig:ii_2}
\end{subfigure}
\begin{subfigure}{0.235\textwidth}
    \includegraphics[scale=0.14]{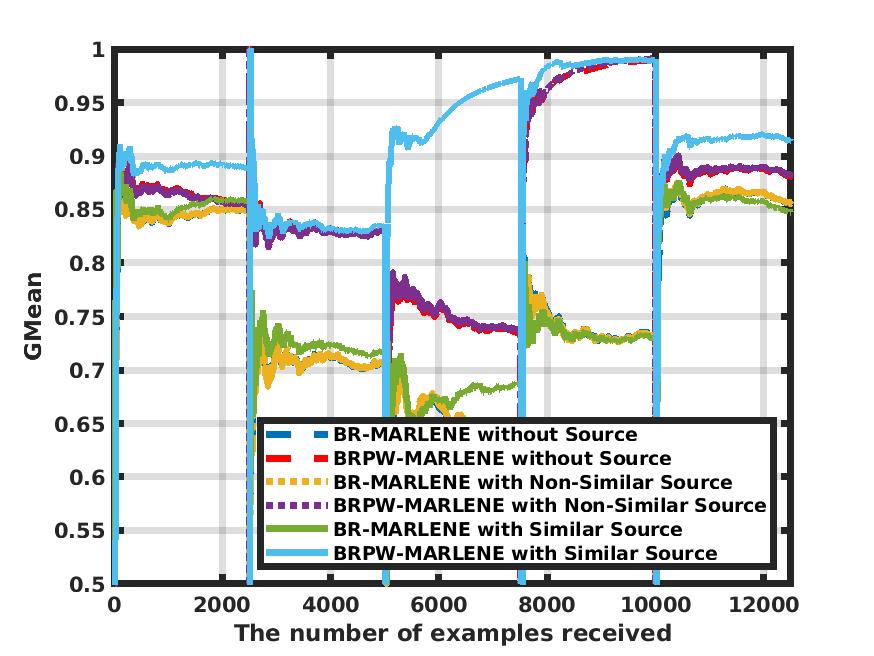}
    \caption{IA; $l_{T,1}$; size of 500}
    \label{subfig:ia_1}
\end{subfigure}
\begin{subfigure}{0.235\textwidth}
    \includegraphics[scale=0.14]{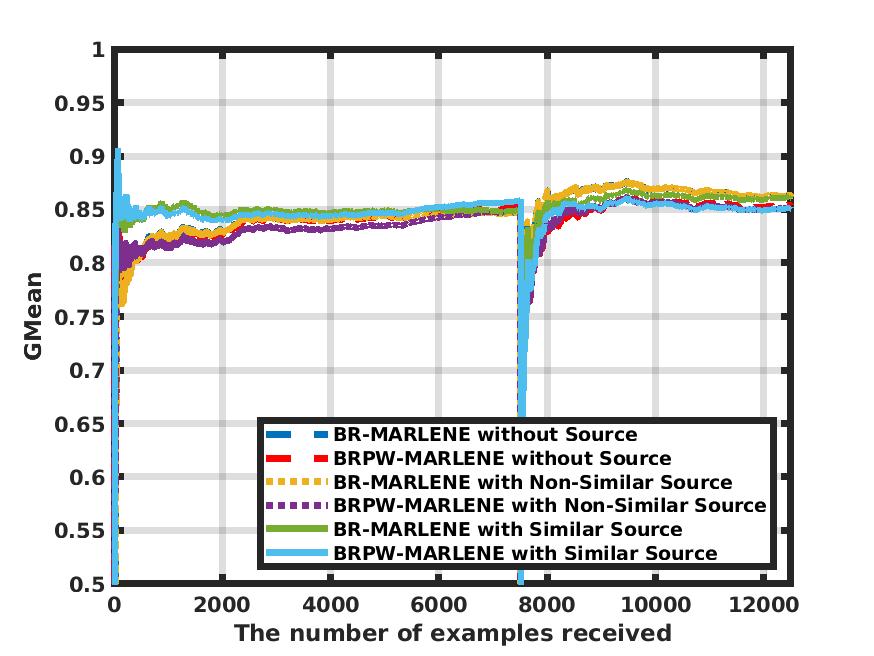}
    \caption{IA; $l_{T,2}$; size of 500}
    \label{subfig:ia_2}
\end{subfigure}

\vspace{1ex}

\begin{subfigure}{0.235\textwidth}
    \includegraphics[scale=0.14]{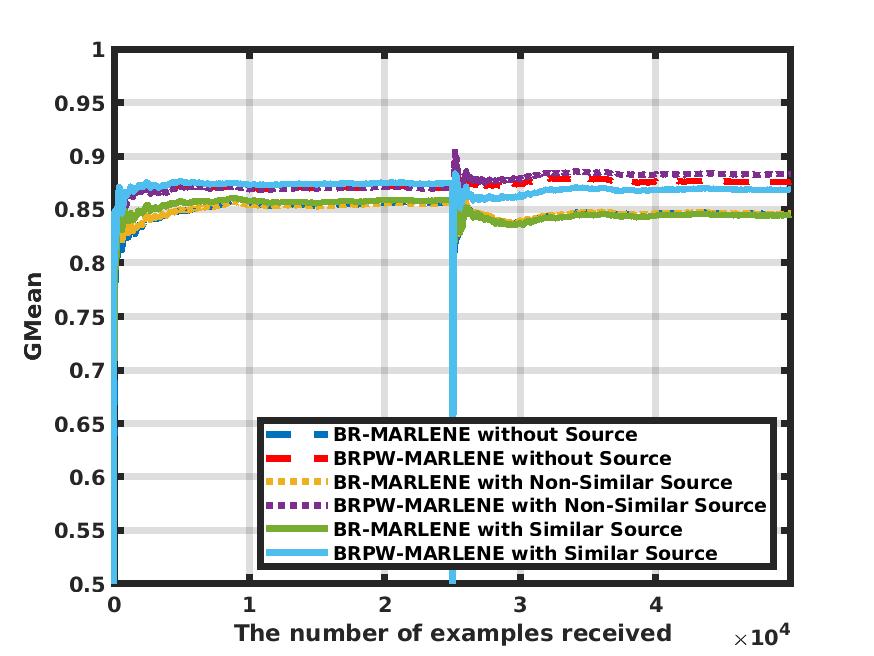}
    \caption{AA; $l_{T,1}$; size of 5000}
    \label{subfig:aa_1}
\end{subfigure}
\begin{subfigure}{0.235\textwidth}
    \includegraphics[scale=0.14]{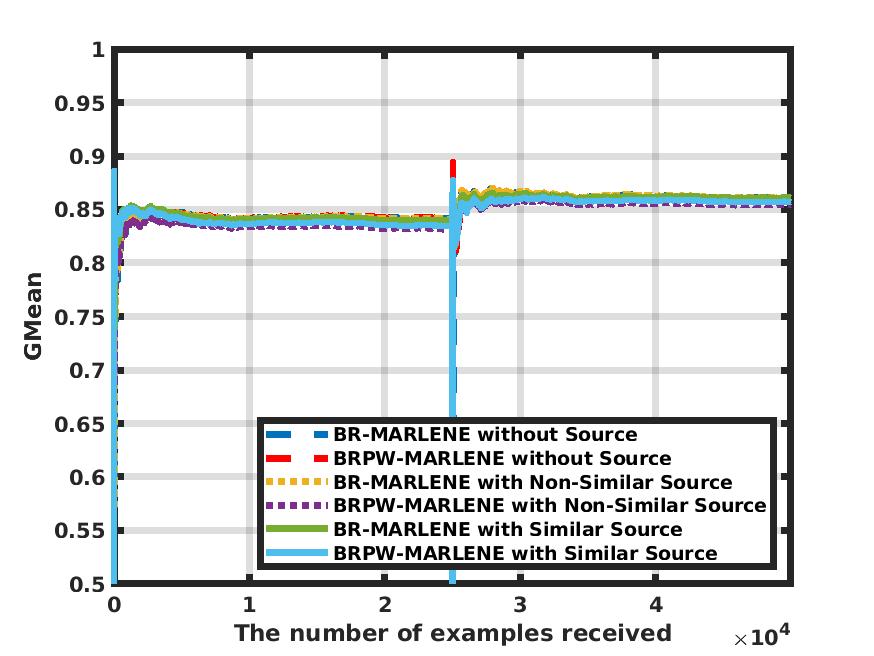}
    \caption{AA; $l_{T,2}$; size of 5000}
    \label{subfig:aa_2}
\end{subfigure}
\begin{subfigure}{0.235\textwidth}
    \includegraphics[scale=0.14]{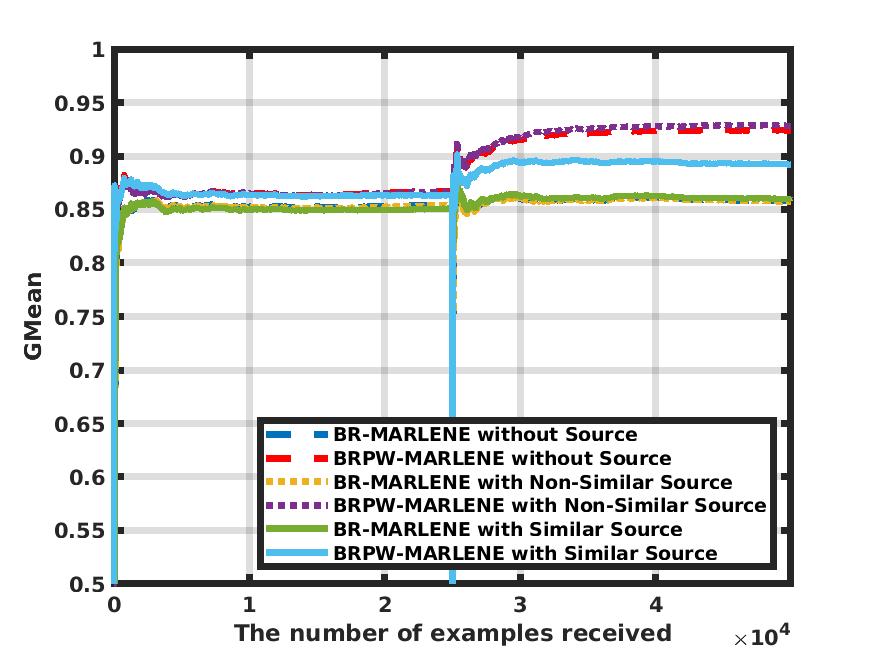}
    \caption{AS; $l_{T,1}$; size of 5000}
    \label{subfig:as_1}
\end{subfigure}
\begin{subfigure}{0.235\textwidth}
    \includegraphics[scale=0.14]{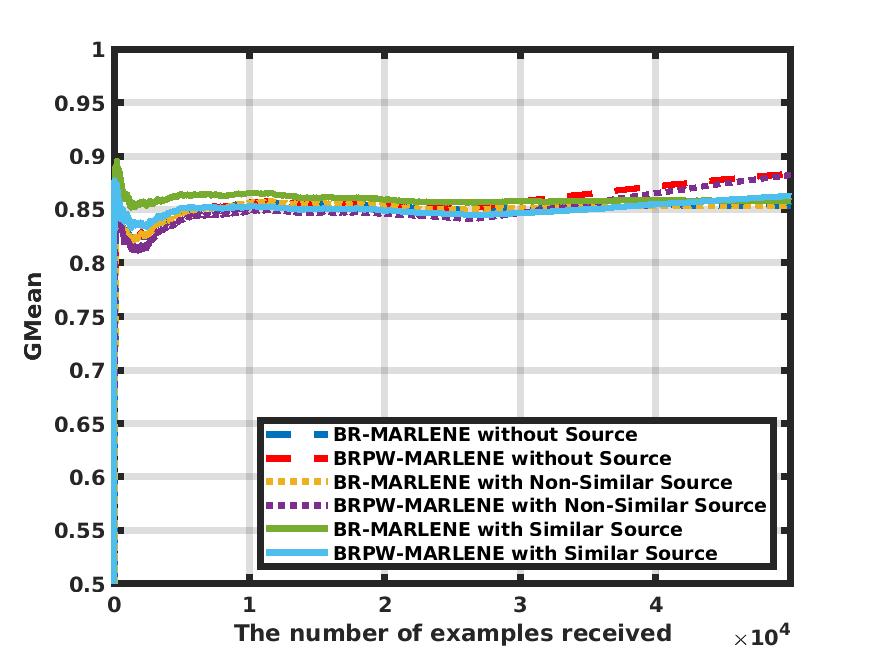}
    \caption{AS; $l_{T,2}$; size of 5000}
    \label{subfig:as_2}
\end{subfigure}

\caption{Average G-Mean of label $l_{T,1}$ and $l_{T,2}$ on Synthetic Datasets.}
\label{fig: mml_marlene_synthetic}
\end{figure*}

To further analyse the performance on each label under different types of concept drift, we plot G-Mean curves against
different labels. G-Mean on each label is calculated prequentially over 30 runs and reset at each drift. This isolates performance on each concept. Representative results are shown in Figure \ref{fig: mml_marlene_synthetic}.

If the concept is stable, similar sources offer the greatest benefit at the beginning of learning (e.g., \ref{subfig:ss_1}, \ref{subfig:ss_2}, \ref{subfig:is_2}, \ref{subfig:ia_2}, \ref{subfig:as_2}) when target examples are limited, but as more target data becomes available, the performance of methods with similar, non-similar, or no source becomes similar. When concept drifts occur frequently, similar sources continue to provide an advantage throughout (e.g., \ref{subfig:is_1}, \ref{subfig:ii_1}, \ref{subfig:ii_2}, \ref{subfig:ia_2}), with greater benefit as drift frequency increases. This is reasonable, as transfer from similar sources is less useful once a large number of target examples is available for training.

\subsection{Effect Analysis of the Transfer Learning}
The strong results of BR/BRPW-MARLENE compared to existing methods (Sections \ref{sec: Results on Real-World Datasets} and \ref{sec: Results on Synthetic Datasets}) demonstrate the effectiveness of our approach. However, it remains to be verified whether this improvement is truly due to our designed transfer mechanisms. This section provides an analysis of whether this is the case.

To assess the role of transfer learning in our methods, we select two datasets (AA with similar sources, 5000 samples; and Yeast) and plot the average weight ratios of all source sub-classifiers (i.e., all sub-classifiers trained on source streams or on other labels in the target stream, excluding the given target label) over 30 runs. For BR-MARLENE, the source sub-classifier weight ratio ($SWR^q$) is the percentage of total weights assigned to source sub-classifiers in the ensemble for label $l_{T,q}$. The average source weight ratio ($ASWR$) is then calculated as:
\begin{align}
    SWR^q = \frac{\sum_{h \in \mathcal{H}_{src}^{BR}} \alpha_h^q}{\sum_{h \in \mathcal{H}^{BR}} \alpha_h^q}, \quad
    ASWR = \frac{\sum_{q=1}^{|\mathcal{L}_T|}SWR^q}{|\mathcal{L}_T|}
\end{align}
where $\mathcal{H}_{src}$ is the set of all source sub-classifiers. For BRPW-MARLENE, the average weight ratio of source PW-classifiers is calculated as:
\begin{align}
    SWR^{q,q'} &= \frac{\sum_{h\in \mathcal{H}_{src}^{PW}} \alpha_h^{q,q'}}{\sum_{h\in \mathcal{H}^{PW}} \alpha_h^{q,q'}}\\
    ASWR &= \frac{\sum_{q=1}^{|\mathcal{L}_T|}\sum_{q'=1, q' \neq q}^{|\mathcal{L}_T|}SWR^{q,q'}}{|\mathcal{L}_T||\mathcal{L}_T|-1}
\end{align}

\begin{figure}[htbp]
\centering
\begin{subfigure}{0.235\textwidth}
    \centering
    \includegraphics[scale=0.2]{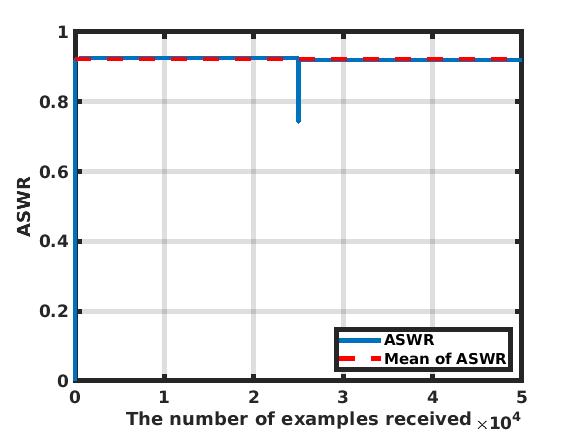}
    \caption{BR-MARLENE}
    \label{subfig:br-marlene_aa_aswr}
\end{subfigure}
\begin{subfigure}{0.235\textwidth}
    \centering
    \includegraphics[scale=0.2]{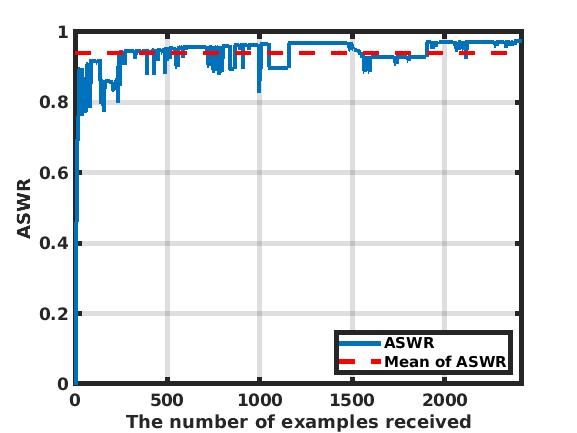}
    \caption{BR-MARLENE}
    \label{subfig:br-marlene_yeast_aswr}
\end{subfigure}
\vspace{1ex}
\begin{subfigure}{0.235\textwidth}
    \centering
    \includegraphics[scale=0.2]{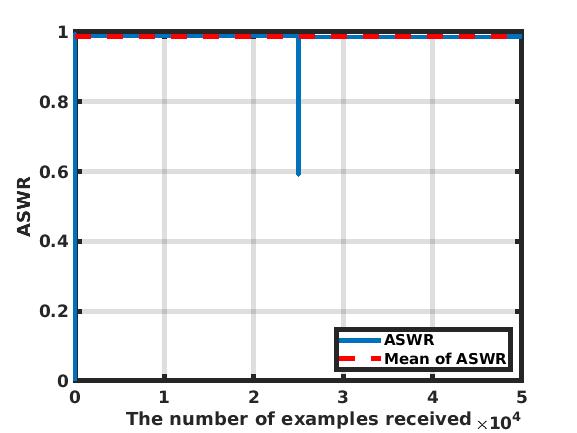}
    \caption{BRPW-MARLENE}
    \label{subfig:brpw-marlene_aa_aswr}
\end{subfigure}
\begin{subfigure}{0.235\textwidth}
    \centering
    \includegraphics[scale=0.2]{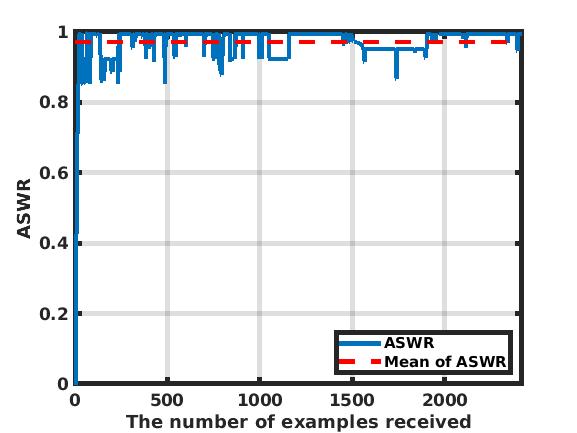}
    \caption{BRPW-MARLENE}
    \label{subfig:brpw-marlene_yeast_aswr}
\end{subfigure}
\caption{
Average Source sub-classifiers' Weight Ratio. 
(\subref{subfig:br-marlene_aa_aswr}) and (\subref{subfig:brpw-marlene_aa_aswr}) show results on AA (size 5000, Similar Source); 
(\subref{subfig:br-marlene_yeast_aswr}) and (\subref{subfig:brpw-marlene_yeast_aswr}) show results on Yeast.
}
\label{fig: marlene_aswr}
\end{figure}

Figure \ref{fig: marlene_aswr} shows that the $ASWR$ of both BR- and BRPW-MARLENE remains high, indicating a strong contribution from source sub-classifiers over time. For the synthetic AA dataset, this weight reflects 
\begin{inparaenum}[1)]
    \item transfer from source data streams to the target stream and 
    \item transfer between different labels within the target stream.
\end{inparaenum}
For Yeast, it reflects inter-label transfer. The consistently high ASWR values confirm that both source-to-target and inter-label transfer are beneficial.

The source PW-classifiers in BRPW-MARLENE have a higher $ASWR$ than source BR-classifiers in BR-MARLENE, since BRPW-MARLENE's ensemble contains many more sub-classifiers; even with some low-weight source sub-classifiers, the total source weight remains high. Spikes and drops in $ASWR$ for both methods on the real-world dataset (Figures \ref{subfig:br-marlene_yeast_aswr}, \ref{subfig:brpw-marlene_yeast_aswr}) suggest that the weighting mechanism may be affected by noise.

\section{Conclusion}
In this paper, we introduced BR-MARLENE, the first approach to transfer knowledge across different labels to boost multi-label classification performance in non-stationary environments. Building on this, we proposed BRPW-MARLENE, a novel extension that further improves classification performance by modelling pairwise label dependencies and transferring knowledge between these dependencies. We conduct comprehensive experiments on real-world and synthetic datasets. The results demonstrate both the effectiveness and clear advantage of these transfer mechanisms.

Although BRPW-MARLENE delivers higher predictive performance, its pairwise dependency modelling is not scalable for problems with a large number of labels. Making BRPW-MARLENE more efficient for large-scale data is a key challenge. Other future directions include experimenting with alternative base classifiers, drift detectors and additional datasets.

\bibliographystyle{IEEEtran}
\bibliography{ref}

\begin{thebibliography}{10}
\providecommand{\url}[1]{#1}
\csname url@samestyle\endcsname
\providecommand{\newblock}{\relax}
\providecommand{\bibinfo}[2]{#2}
\providecommand{\BIBentrySTDinterwordspacing}{\spaceskip=0pt\relax}
\providecommand{\BIBentryALTinterwordstretchfactor}{4}
\providecommand{\BIBentryALTinterwordspacing}{\spaceskip=\fontdimen2\font plus
\BIBentryALTinterwordstretchfactor\fontdimen3\font minus \fontdimen4\font\relax}
\providecommand{\BIBforeignlanguage}[2]{{%
\expandafter\ifx\csname l@#1\endcsname\relax
\typeout{** WARNING: IEEEtran.bst: No hyphenation pattern has been}%
\typeout{** loaded for the language `#1'. Using the pattern for}%
\typeout{** the default language instead.}%
\else
\language=\csname l@#1\endcsname
\fi
#2}}
\providecommand{\BIBdecl}{\relax}
\BIBdecl

\bibitem{read2022learning}
J.~Read and I.~{\v{Z}}liobait{\.e}, ``Learning from data streams: An overview and update,'' \emph{arXiv preprint arXiv:2212.14720}, 2022.

\bibitem{spyromitros2011dealing}
E.~Spyromitros-Xioufis, M.~Spiliopoulou, G.~Tsoumakas, and I.~Vlahavas, ``Dealing with concept drift and class imbalance in multi-label stream classification,'' \emph{Department of Computer Science, Aristotle University of Thessaloniki}, 2011.

\bibitem{buyukccakir2018novel}
A.~B{\"u}y{\"u}k{\c{c}}akir, H.~Bonab, and F.~Can, ``A novel online stacked ensemble for multi-label stream classification,'' in \emph{Proceedings of the 27th ACM International Conference on Information and Knowledge Management}, 2018, pp. 1063--1072.

\bibitem{zhuang2020comprehensive}
F.~Zhuang, Z.~Qi, K.~Duan, D.~Xi, Y.~Zhu, H.~Zhu, H.~Xiong, and Q.~He, ``A comprehensive survey on transfer learning,'' \emph{Proceedings of the IEEE}, vol. 109, no.~1, pp. 43--76, 2020.

\bibitem{kubat1997addressing}
M.~Kubat, S.~Matwin \emph{et~al.}, ``Addressing the curse of imbalanced training sets: one-sided selection,'' in \emph{14th International Conference on Machine Learning}, vol.~97.\hskip 1em plus 0.5em minus 0.4em\relax Citeseer, 1997, pp. 79--86.

\bibitem{minku2012ddd}
L.~L. Minku and X.~Yao, ``Ddd: A new ensemble approach for dealing with concept drift,'' \emph{IEEE transactions on knowledge and data engineering}, vol.~24, no.~4, pp. 619--633, 2011.

\bibitem{zhao2014online}
P.~Zhao, S.~C. Hoi, J.~Wang, and B.~Li, ``Online transfer learning,'' \emph{Artificial Intelligence}, vol. 216, pp. 76--102, 2014.

\bibitem{du2019multi}
H.~Du, L.~L. Minku, and H.~Zhou, ``Multi-source transfer learning for non-stationary environments,'' in \emph{2019 International Joint Conference on Neural Networks (IJCNN)}.\hskip 1em plus 0.5em minus 0.4em\relax IEEE, 2019, pp. 1--8.

\bibitem{yu2024online}
E.~Yu, J.~Lu, B.~Zhang, and G.~Zhang, ``Online boosting adaptive learning under concept drift for multistream classification,'' in \emph{{AAAI} Conference on Artificial Intelligence}, vol.~38, no.~15, 2024, pp. 16\,522--16\,530.

\bibitem{du2020marline}
H.~Du, L.~L. Minku, and H.~Zhou, ``Marline: Multi-source mapping transfer learning for non-stationary environments,'' in \emph{IEEE International Conference on Data Mining}.\hskip 1em plus 0.5em minus 0.4em\relax IEEE, 2020, pp. 122--131.

\bibitem{sun2018concept}
Y.~Sun, K.~Tang, Z.~Zhu, and X.~Yao, ``Concept drift adaptation by exploiting historical knowledge,'' \emph{IEEE Transactions on Neural Networks and Learning Systems}, vol.~29, no.~10, pp. 4822--4832, 2018.

\bibitem{yang2021concept}
C.~Yang, Y.-m. Cheung, J.~Ding, and K.~C. Tan, ``Concept drift-tolerant transfer learning in dynamic environments,'' \emph{IEEE Transactions on Neural Networks and Learning Systems}, vol.~33, no.~8, pp. 3857--3871, 2021.

\bibitem{jiao2025otl}
B.~Jiao and S.~Liu, ``Otl-ce: Online transfer learning for data streams with class evolution,'' \emph{Neurocomputing}, p. 129470, 2025.

\bibitem{renchunzi2022automatic}
X.~Renchunzi and M.~Pratama, ``Automatic online multi-source domain adaptation,'' \emph{Information Sciences}, vol. 582, pp. 480--494, 2022.

\bibitem{godbole2004discriminative}
S.~Godbole and S.~Sarawagi, ``Discriminative methods for multi-labeled classification,'' in \emph{Pacific-Asia Conference on Knowledge Discovery and Data Mining}.\hskip 1em plus 0.5em minus 0.4em\relax Springer, 2004, pp. 22--30.

\bibitem{read2009classifier}
J.~Read, B.~Pfahringer, G.~Holmes, and E.~Frank, ``Classifier chains for multi-label classification,'' in \emph{Joint European Conference on Machine Learning and Knowledge Discovery in Databases}.\hskip 1em plus 0.5em minus 0.4em\relax Springer, 2009, pp. 254--269.

\bibitem{furnkranz2008multilabel}
J.~F{\"u}rnkranz, E.~H{\"u}llermeier, E.~L. Menc{\'\i}a, and K.~Brinker, ``Multilabel classification via calibrated label ranking,'' \emph{Machine Learning}, vol.~73, no.~2, pp. 133--153, 2008.

\bibitem{read2008multi}
J.~Read, B.~Pfahringer, and G.~Holmes, ``Multi-label classification using ensembles of pruned sets,'' in \emph{2008 eighth IEEE International Conference on Data Mining}.\hskip 1em plus 0.5em minus 0.4em\relax IEEE, 2008, pp. 995--1000.

\bibitem{osojnik2017multi}
A.~Osojnik, P.~Panov, and S.~D{\v{z}}eroski, ``Multi-label classification via multi-target regression on data streams,'' \emph{Machine Learning}, vol. 106, no.~6, pp. 745--770, 2017.

\bibitem{wu2023weighted}
H.~Wu, M.~Han, Z.~Chen, M.~Li, and X.~Zhang, ``A weighted ensemble classification algorithm based on nearest neighbors for multi-label data stream,'' \emph{ACM Transactions on Knowledge Discovery from Data}, vol.~17, no.~5, pp. 1--21, 2023.

\bibitem{zou2024weak}
Y.~Zou, X.~Hu, P.~Li, and J.~Hu, ``Weak multi-label data stream classification under distribution changes in labels,'' \emph{IEEE Transactions on Big Data}, 2024.

\bibitem{li2022high}
P.~Li, H.~Zhang, X.~Hu, and X.~Wu, ``High-dimensional multi-label data stream classification with concept drifting detection,'' \emph{IEEE Transactions on Knowledge and Data Engineering}, vol.~35, no.~8, pp. 8085--8099, 2022.

\bibitem{esteban2024hoeffding}
A.~Esteban, A.~Cano, A.~Zafra, and S.~Ventura, ``Hoeffding adaptive trees for multi-label classification on data streams,'' \emph{Knowledge-Based Systems}, vol. 304, p. 112561, 2024.

\bibitem{bifet2010leveraging}
A.~Bifet, G.~Holmes, and B.~Pfahringer, ``Leveraging bagging for evolving data streams,'' in \emph{Joint European Conference on Machine Learning and Knowledge Discovery in Databases}, 2010, pp. 135--150.

\bibitem{read2012scalable}
J.~Read, A.~Bifet, G.~Holmes, and B.~Pfahringer, ``Scalable and efficient multi-label classification for evolving data streams,'' \emph{Machine Learning}, vol.~88, no. 1-2, pp. 243--272, 2012.

\bibitem{domingos2000mining}
P.~Domingos and G.~Hulten, ``Mining high-speed data streams,'' in \emph{Proceedings of the sixth ACM SIGKDD International Conference on Knowledge Discovery and Data Mining}, 2000, pp. 71--80.

\bibitem{wang2013concept}
S.~Wang, L.~L. Minku, D.~Ghezzi, D.~Caltabiano, P.~Tino, and X.~Yao, ``Concept drift detection for online class imbalance learning,'' in \emph{The 2013 International Joint Conference on Neural Networks (IJCNN)}.\hskip 1em plus 0.5em minus 0.4em\relax IEEE, 2013, pp. 1--10.

\bibitem{you2021online}
D.~You, Y.~Wang, J.~Xiao, Y.~Lin, M.~Pan, Z.~Chen, L.~Shen, and X.~Wu, ``Online multi-label streaming feature selection with label correlation,'' \emph{IEEE Transactions on Knowledge and Data Engineering}, 2021.

\bibitem{sousa2018multi}
R.~Sousa and J.~Gama, ``Multi-label classification from high-speed data streams with adaptive model rules and random rules,'' \emph{Progress in Artificial Intelligence}, vol.~7, no.~3, pp. 177--187, 2018.

\bibitem{vergara2025multi}
M.~Vergara, B.~Bustos, I.~Sipiran, T.~Schreck, and S.~Lengauer, ``Multi-label learning on low label density sets with few examples,'' \emph{Expert Systems with Applications}, vol. 265, p. 125942, 2025.

\bibitem{zheng2019survey}
X.~Zheng, P.~Li, Z.~Chu, and X.~Hu, ``A survey on multi-label data stream classification,'' \emph{IEEE Access}, vol.~8, pp. 1249--1275, 2019.

\bibitem{bifet2010moa}
A.~Bifet, G.~Holmes, R.~Kirkby, and B.~Pfahringer, ``Moa: Massive online analysis,'' \emph{The Journal of Machine Learning Research}, vol.~11, no. May, pp. 1601--1604, 2010.

\bibitem{read2016meka}
J.~Read, P.~Reutemann, B.~Pfahringer, and G.~Holmes, ``Meka: a multi-label/multi-target extension to weka,'' in \emph{The Journal of Machine Learning Research}, 2016.

\bibitem{bifet2009new}
A.~Bifet, G.~Holmes, B.~Pfahringer, R.~Kirkby, and R.~Gavalda, ``New ensemble methods for evolving data streams,'' in \emph{Proceedings of the 15th ACM SIGKDD International Conference on Knowledge Discovery and Data mining}, 2009, pp. 139--148.

\bibitem{gama2014survey}
J.~Gama, I.~{\v{Z}}liobait{\.e}, A.~Bifet, M.~Pechenizkiy, and A.~Bouchachia, ``A survey on concept drift adaptation,'' \emph{ACM Computing Surveys (CSUR)}, vol.~46, no.~4, p.~44, 2014.

\bibitem{wu2020multi}
G.~Wu and J.~Zhu, ``Multi-label classification: do hamming loss and subset accuracy really conflict with each other?'' \emph{Advances in Neural Information Processing Systems}, vol.~33, pp. 3130--3140, 2020.

\bibitem{heydarian2022mlcm}
M.~Heydarian, T.~E. Doyle, and R.~Samavi, ``Mlcm: Multi-label confusion matrix,'' \emph{Ieee Access}, vol.~10, pp. 19\,083--19\,095, 2022.

\bibitem{charte2015addressing}
F.~Charte, A.~J. Rivera, M.~J. del Jesus, and F.~Herrera, ``Addressing imbalance in multilabel classification: Measures and random resampling algorithms,'' \emph{Neurocomputing}, vol. 163, pp. 3--16, 2015.

\bibitem{wang2014resampling}
S.~Wang, L.~L. Minku, and X.~Yao, ``Resampling-based ensemble methods for online class imbalance learning,'' \emph{IEEE Transactions on Knowledge and Data Engineering}, vol.~27, no.~5, pp. 1356--1368, 2014.

\end{thebibliography}

\end{document}